\pdfoutput=1

\documentclass[11pt]{article}

\usepackage[final]{acl}
\usepackage[T1]{fontenc}
\usepackage{CJKutf8}
\usepackage[utf8]{inputenc}
\newcommand{\zh}[1]{\begin{CJK}{UTF8}{gbsn}#1\end{CJK}}
\usepackage{times}
\usepackage{latexsym}
\usepackage{microtype}
\usepackage{inconsolata}
\usepackage{graphicx}
\definecolor{gold}{RGB}{234, 51, 35}

\usepackage{arydshln}
\usepackage[most]{tcolorbox}
\usepackage{multirow}
\newtcolorbox{taskbox}[2][]{
	enhanced, breakable,
	colframe=blue3!40,
	colback=blue5!5,
	arc=1mm,
	outer arc=1mm,
	fontupper=\small,
	fontlower=\small,
	coltitle=blue1,
	fonttitle=\bfseries,
	boxsep=1mm,
	left=0mm,
	right=0mm,
	top=0mm,
	bottom=0mm,
	before={\noindent},
	segmentation style={solid, blue3},
	title=#2,
	#1
}
\newtcolorbox{mybox}[2][]{
	width=\columnwidth,
	colback = gray!8, 
	colframe = gray!8, 
	boxsep=0pt,left=0pt,right=8pt,top=8pt,bottom=8pt,
	fontupper=\linespread{0.9}\selectfont,
	title=#2,#1}
\usepackage{amsmath}
\usepackage{mathtools}
\usepackage{amsfonts}
\usepackage{amssymb}
\usepackage{pifont}
\usepackage{tabularx}
\usepackage{arydshln}
\usepackage{adjustbox}
\usepackage{paralist} 
\usepackage{booktabs} 
\usepackage{subfig}
\usepackage{xcolor}
\usepackage{colortbl}
\usepackage{booktabs}
\usepackage{adjustbox}
\usepackage{arydshln}
\usepackage{tikz}
\usepackage[edges]{forest}
\usepackage{dsfont}
\usepackage{courier}
\usepackage{pgfplots}
\usepackage{makecell}
\usepackage[most]{tcolorbox}
\definecolor{darkolivegreen}{rgb}{0.33, 0.42, 0.18}
\definecolor{my_green}{RGB}{40,154,121}
\definecolor{my_yellow}{RGB}{255,165,0}
\definecolor{my_red}{RGB}{176,46,46}

\usepackage{hyperref}

\definecolor{red}{RGB}{184,49,55}

\definecolor{blue}{RGB}{55,83,156}

\definecolor{green}{RGB}{100,141,63}

\newcommand{\correctmark}{\textcolor{my_green}{\ding{52}}} 
\newcommand{\errormark}{\textcolor{my_red}{\ding{56}}}

\definecolor{myblue}{RGB}{215,226,240}
\definecolor{mygreen}{RGB}{229,238,226}

\usepackage[T1]{fontenc}

\usepackage[utf8]{inputenc}

\usepackage{microtype}

\usepackage{inconsolata}

\usepackage{graphicx}

\title{\textit{CCHall}: 
A Novel Benchmark for Joint Cross-Lingual and \\ Cross-Modal Hallucinations Detection in Large Language Models}

\author{
	Yongheng Zhang$^{1,2*}$~~
	Xu Liu$^{1}$\thanks{~~Equal Contribution. $^\dagger$ Corresponding Author.}~~
	\textbf{Ruoxi Zhou}$^{1}$~~
	\textbf{Qiguang Chen}$^{1}$
	\\
	\textbf{Hao Fei}$^{3}$~~
	\textbf{Wenpeng Lu}$^{4}$~~
    \textbf{Libo Qin}$^{1,2\dagger}$\\
	$^{1}$School of Computer Science and Engineering, Central South University, China\\
	$^{2}$Key Laboratory of Data Intelligence and Advanced Computing in Provincial Universities, \\ $^{2}$Soochow University, China 
	$^{3}$National University of Singapore, Singapore\\
	$^{4}$Key Laboratory of Computing Power Network and Information Security, Ministry of Education\\
	$^{4}$Qilu University of Technology (Shandong Academy of Sciences), China\\
}

\begin{document}
\maketitle
\begin{abstract}
Investigating hallucination issues in large language models (LLMs) within cross-lingual and cross-modal scenarios can greatly advance the 
large-scale deployment in real-world applications. Nevertheless, the current studies are limited to a single scenario, either cross-lingual or cross-modal, leaving a gap in the exploration of hallucinations in the joint cross-lingual and cross-modal scenarios. 
Motivated by this, we introduce a novel joint \textbf{C}ross-lingual and \textbf{C}ross-modal \textbf{Hall}ucinations benchmark (\texttt{CCHall}) to fill this gap.
Specifically, \texttt{CCHall} simultaneously incorporates both cross-lingual and cross-modal hallucination scenarios, which can be used to assess the cross-lingual and cross-modal capabilities of LLMs. Furthermore, we conduct a comprehensive evaluation on \texttt{CCHall}, exploring both mainstream open-source and closed-source LLMs. The experimental results highlight that current LLMs still struggle with \texttt{CCHall}. We hope \texttt{CCHall} can serve as a valuable resource to assess LLMs in joint cross-lingual and cross-modal scenarios.

\end{abstract}

\vspace{1mm}
\section{Introduction}

Large language models (LLMs) have made significant progress in recent years \cite{meta2024llama, qin2024large, wang2025comprehensive}, driving remarkable advancements across a wide range of diverse fields and applications \cite{wei2022chain, zhang2025multi}. 
However, it is disheartening that the issue of hallucinations in LLMs remains unresolved, significantly hindering their large-scale deployment in real-world applications \cite{park2024mitigating, yu2025survey}. This phenomenon is particularly severe in both cross-lingual and cross-modal contexts, potentially causing critical errors in applications such as medical diagnosis, image captioning, and speech-to-text conversion \cite{guerreiro2023hallucinations, regan2024massive, sriramanan2025llm}.

\begin{figure}[t]
  \includegraphics[width=\columnwidth]{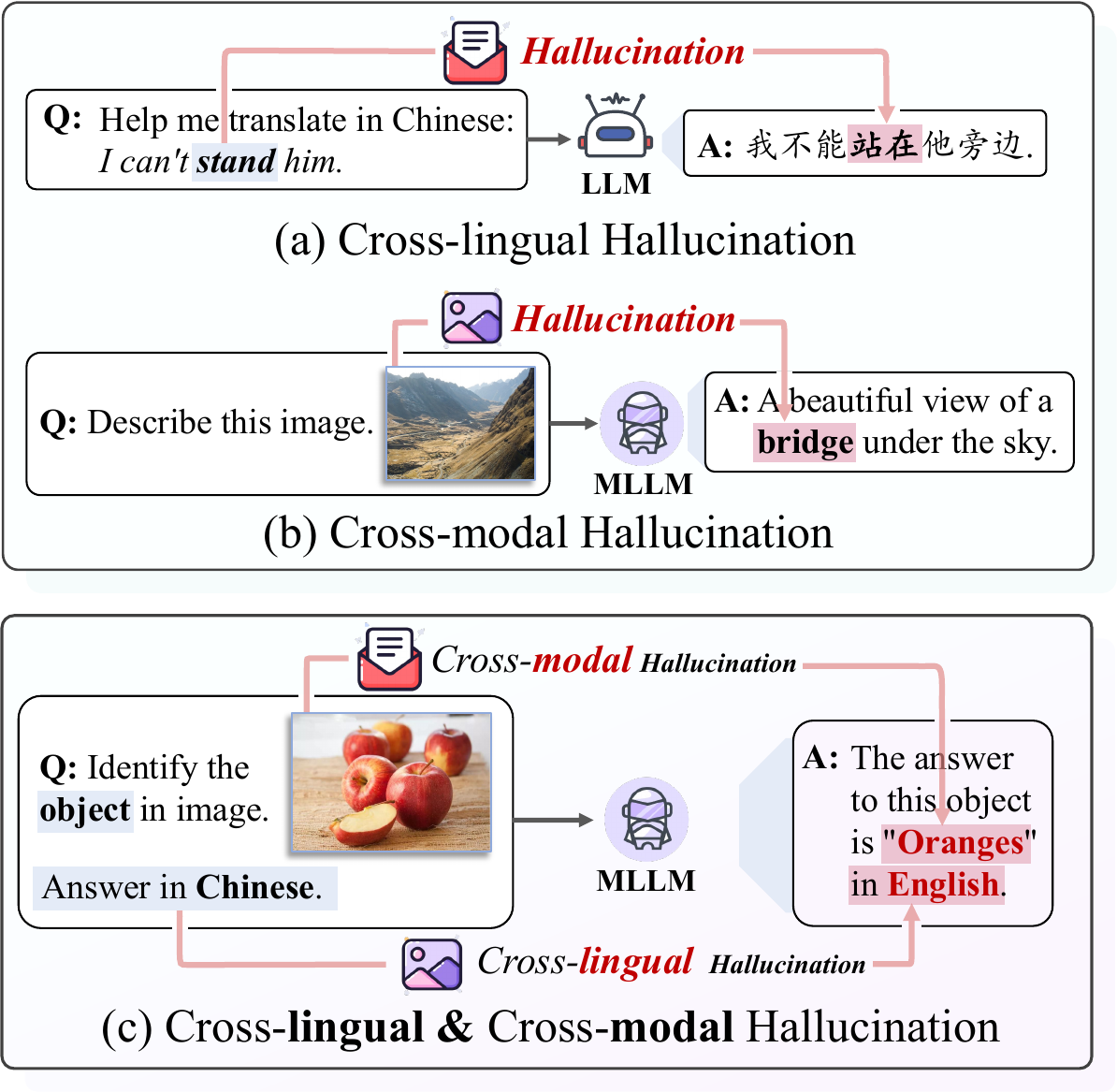}
  \vspace{-6mm}
  \caption{(a) Cross-lingual hallucination: A cross-lingual hallucination emerged: the erroneous translation of ``stand'' as ``\zh{站在}''. Here it should be ``\zh{忍受}''; (b) Cross-modal hallucination: A cross-modal hallucination occurred, fabricating a ``bridge''; (c) Cross-lingual and Cross-modal hallucination: A cross-modal hallucination fabricated ``Oranges'' and a cross-lingual hallucination did not use Chinese in its Answer.
  }
  \vspace{-3mm}
  \label{intro}
\end{figure}

\begin{figure*}[t]
	\centering 
	\begin{minipage}{0.41\textwidth} 
		\centering
		\vspace{-1mm}
		\includegraphics[width=\linewidth]{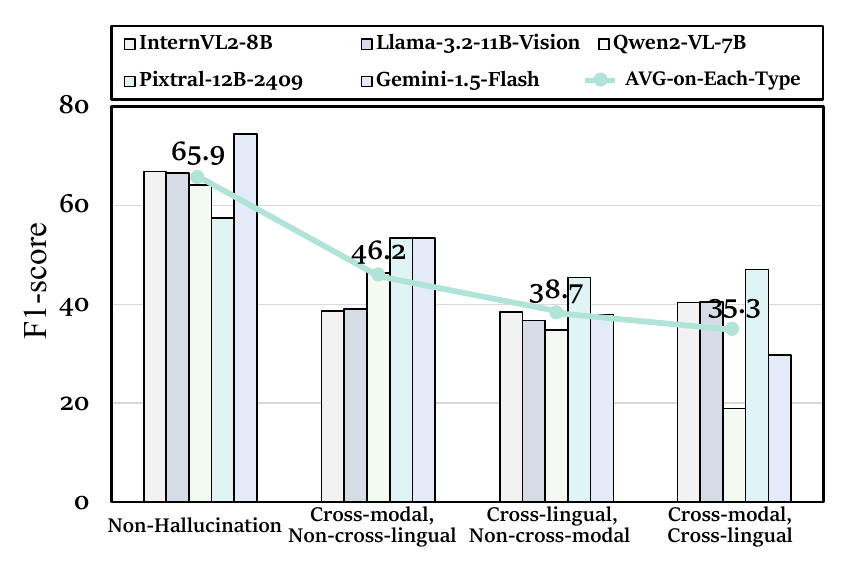}
		\vspace{-7mm}
		\caption*{(a) F1-score on different hallucination types.}
	\end{minipage}
	\hspace{0\textwidth} 
	\begin{minipage}{0.56\textwidth} 
		\begin{adjustbox}{width=\textwidth}
		\begin{tabular}{{l}*{6}{c}}
			\toprule	
			 \multirow{2}{*}{\vspace{-2mm}\textbf{Benchmark}} & \multicolumn{3}{c}{\textbf{Methods}} \\ 
			\cmidrule{2-4}
			& \textbf{Cross-lingual} &  \textbf{Cross-modal} & \textbf{Cross-(Lingual \& Modal)} \\ 
			\midrule
			\rowcolor[rgb]{ .970, .978, .999 }
			XL-Sum~\cite{hasan2021xlsumlargescalemultilingualabstractive} & \correctmark & \errormark & \errormark  \\
  			\rowcolor[rgb]{ .970, .978, .999 }
			X-fact~\cite{gupta2021x} & \correctmark & \errormark & \errormark \\
  			\rowcolor[rgb]{ .970, .978, .999 }
			\text{HalOmi} \cite{dale2023halomimanuallyannotatedbenchmark} & \correctmark & \errormark & \errormark \\
   			\rowcolor[rgb]{ .970, .978, .999 }
			\text{MM-Eval}~\cite{son2024mmevalmultilingualmetaevaluationbenchmark} & \correctmark & \errormark & \errormark  \\
  			\rowcolor[rgb]{ .970, .978, .999 }
			XTRUST~\cite{li2024xtrustmultilingualtrustworthinesslarge} & \correctmark & \errormark & \errormark  \\
			\midrule
 			\rowcolor[rgb]{ .970, .978, .999 }
			\text{CHAIR}~\cite{rohrbach2018object} & \errormark & \correctmark & \errormark \\
 			\rowcolor[rgb]{ .970, .978, .999 }
			\text{POPE}~\cite{li2023evaluating}  & \errormark & \correctmark & \errormark \\
 			\rowcolor[rgb]{ .970, .978, .999 }
			\text{M-HalDetect}~\cite{gunjal2024detecting} & \errormark & \correctmark & \errormark \\
 			\rowcolor[rgb]{ .970, .978, .999 }
			\text{HallusionBench}~\cite{guan2024hallusionbench} & \errormark & \correctmark & \errormark \\
			\rowcolor[rgb]{ .970, .978, .999 }
			\text{MHaluBench}~\cite{chen2024unified} & \errormark & \correctmark & \errormark \\
			\midrule
			\rowcolor[rgb]{ .951, .985, .970 }
			\texttt{\textbf{CCHall}} (ours) & \correctmark & \correctmark & \correctmark \\
			\bottomrule
		\end{tabular}

		\end{adjustbox}
		\vspace{-2mm}
		\caption*{(b) Comparison of \texttt{CCHall} with related datasets.}
	\end{minipage}
\caption{(a) Fine-grained performance analysis of MLLMs F1-score for different hallucination types in \texttt{CCHall}. The F1-score of MLLMs on joint cross-lingual and cross-modal hallucinations is 3.4 lower than when addressing cross-lingual hallucinations independently and 10.9 lower than when addressing cross-modal hallucinations independently; (b) A comparative analysis of \texttt{CCHall} with related cross-modal and cross-lingual datasets.}
\vspace{-2mm}
\label{data_comparison}
\end{figure*}

Recent research has focused on cross-lingual and cross-modal hallucinations in LLMs \cite{son2024mmevalmultilingualmetaevaluationbenchmark, guan2024hallusionbench, bai2024hallucination}, aiming to understand their causes, including:

\vspace{1mm}

\textbf{(1) Cross-lingual Hallucinations}: As shown in Figure~\ref{intro} (a), Cross-lingual hallucination refers to the failure to follow instructions or the generation of inaccurate outputs when processing different languages. Specifically, \citet{qiu2023detecting} introduce mFACT to evaluate the faithfulness of summaries and observe that LLMs hallucinate more in non-English languages. \citet{dale2023halomi} release an annotated dataset covering 18 translation directions, tackling hallucinations. \citet{herrlein2024anhalten} further extend English hallucination detection to German and apply it in long-context scenarios.

\textbf{(2) Cross-modal Hallucinations}: As shown in Figure~\ref{intro} (b), Cross-modal hallucination refers to the fabrication and inconsistencies that arise in multimodal large language models (MLLMs) when reasoning across different modalities. Specifically, \citet{liu2023hallusionbench} develop an innovative benchmark for image-context reasoning using image-question pairs created by human experts. \citet{leng2024curse} conduct a detailed and systematic study of multimodal hallucinations and introduce the ``Curse of Multi-Modalities'' benchmark for MLLM evaluation. \citet{yan2024evaluating} introduce a Quality Measurement framework, which aims to evaluate the validity of existing hallucination benchmarks, ensuring that they can accurately assess model outputs.

Despite significant advancements, current research on hallucinations in MLLMs remains overly optimistic, as it \textit{\textbf{primarily focuses}} on either cross-lingual or cross-modal scenarios \textit{\textbf{in isolation}}, leaving a gap in the joint cross-lingual and cross-modal context. In fact, compared to addressing cross-lingual or cross-modal scenarios individually, the joint cross-lingual and cross-modal context is more widely applied in real-world scenarios and presents even  \textit{\textbf{greater challenges}}  \cite{qin2023crosslingualpromptingimprovingzeroshot, chen2024m, zhang2024autocap, wang2024multimodal,castillo2025hierarchical}, as language differences and multimodal inputs increase hallucinations in MLLMs. As illustrated in Figure~\ref{intro}~(c), MLLMs must account not only for the alignment between images and text but also for the alignment across multiple language queries. 
As shown in Figure~\ref{data_comparison}~(a), numerous models struggle with cross-lingual and cross-modal hallucinations, exhibiting poor performance. More frustratingly, there is currently no research to investigate this critical issue.

\begin{figure*}[t]
	\centering
	\includegraphics[width=160mm]{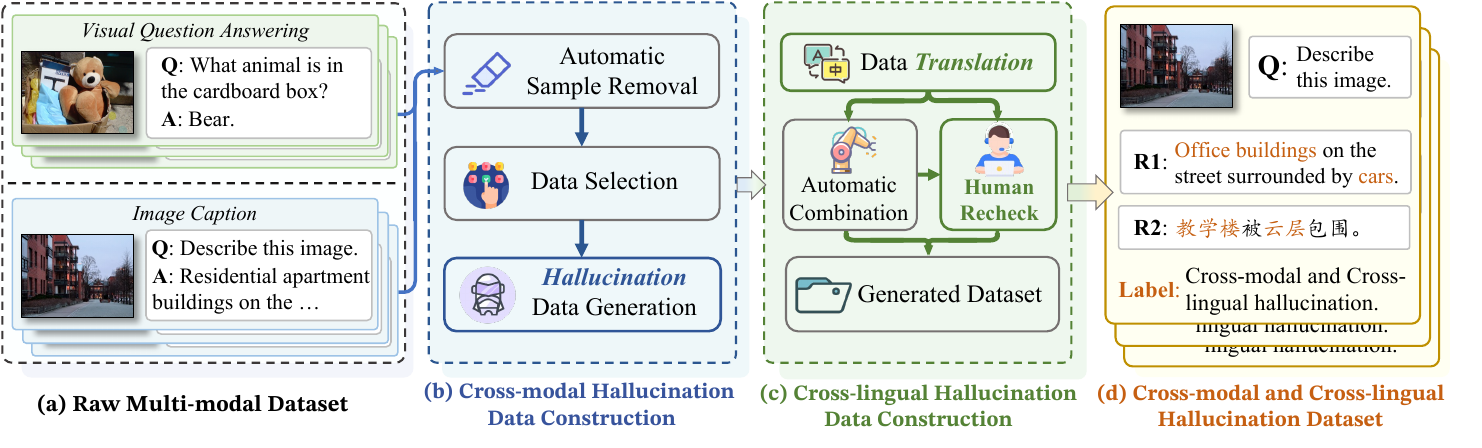}
	\caption{The construction process of \texttt{CCHall} includes: (a) Raw Multi-modal Dataset Selection ($\S \ref{raw}$), (b) Cross-modal Hallucination Data Construction ($\S \ref{Cross-modal Hallucination Data}$), (c) Cross-lingual Hallucination Data Construction ($\S \ref{Cross-lingual Hallucination Data}$), and (d) Cross-modal and Cross-lingual Hallucination Dataset ($\S \ref{CCHall Dataset}$).}
	\label{dataset}
\end{figure*}

Motivated by this, we introduce a novel benchmark, \textbf{C}ross-lingual and \textbf{C}ross-modal \textbf{Hall}ucinations (\texttt{CCHall}), to fill the gap. As shown in Figure~\ref{data_comparison}~(b), \texttt{CCHall} provides an evaluation framework that encompasses \textbf{\textit{not only}} individual cross-lingual and cross-modal hallucinations \textbf{\textit{but also}} joint Cross-lingual and Cross-modal Hallucinations. Furthermore, \texttt{CCHall} covers a wide range of topics and nature scenes, aiming to provide a comprehensive evaluation of MLLMs in cross-lingual and cross-modal hallucination scenarios.
Through evaluation experiments on \texttt{CCHall}, we derive the following \textbf{\textit{key takeaways}}: \textbf{(1)} Current MLLMs are still struggling in \texttt{CCHall} benchmark; \textbf{(2)} In hallucination mitigation methods, basic strategies are more suitable for MLLMs with fewer parameters (<12B), while powerful MLLMs exhibit advantages when using more advanced strategies; \textbf{(3)} The use of multilingual contexts and tool-assisted invocation can effectively mitigate hallucinations in MLLMs.
 
Our contributions to our work are as follows:
\vspace{-2mm}
\begin{itemize}
\item [(1)] We identify and point out that current research on hallucinations tends to be overly optimistic, as it mainly addresses either cross-lingual or cross-modal scenarios separately.
 \vspace{-2mm}
\item [(2)] To address the existing gap in the detection of joint cross-lingual and cross-modal hallucinations, we introduce \textbf{C}ross-lingual and \textbf{C}ross-modal \textbf{Hall}ucinations benchmark (\texttt{CCHall}).
 \vspace{-7mm}
\item [(3)] We evaluate \texttt{CCHall} across a range of MLLMs and diverse scenarios, hoping to provide insights that could help mitigate cross-lingual and cross-modal hallucinations.
\end{itemize}

 \vspace{-2mm}
All data and source code are open-sourced at \url{https://github.com/BRZ911/CCHall}.

\section{Background}

\subsection{Cross-lingual Hallucinations}

Cross-linguistic hallucination \cite{dale2023halomimanuallyannotatedbenchmark} occurs when the LLMs generate an answer $ \mathcal{A}_t $ in the target language $ L_t $ that deviates from the correct answer $ \mathcal{A}_{gold} $. As shown in Figure~\ref{intro} (a), given a question $ Q $ in the source language $ L_s $ and a prompt $ P_{s \rightarrow t} $ to respond in $ L_t $, which is denoted as:

\begin{equation}
	\mathcal{A}_t = \underset{\mathcal{A}}{\operatorname{argmax}} \ P(\mathcal{A}\ |\ Q, P_{s \rightarrow t}),
\end{equation}
where $ P(\mathcal{A}\ |\ Q, P_{s \rightarrow t}) $ is the probability of generating answer $ \mathcal{A} $ given $ Q $ and $ P_{s \rightarrow t} $. Hallucination occurs when multilingual instructions are not followed or when incorrect reasoning is provided.

\subsection{Cross-modal Hallucinations}

Cross-modal hallucination \cite{rohrbach2018object} occurs when the LLMs generate a textual answer $ \mathcal{A}_t $ that does not accurately reflect the content of an input image $ \mathcal{A}_{gold} $. As shown in Figure~\ref{intro} (b), given an image $ I $, a question $ Q $, and a prompt $ P $, the model aims to output an answer $ \mathcal{A} $ that aligns with both $ I $ and $ Q $, which is denoted as:

\begin{equation}
    \mathcal{A}_t = \underset{\mathcal{A}}{\operatorname{argmax}} \ P(\mathcal{A}\ |\ Q, I, P),
\end{equation}
where $ P(\mathcal{A}\ |\ Q, I, P) $ represents the probability of generating answer $ \mathcal{A} $ given $ I $, $ Q $, and $ P $. Cross-modal hallucination occurs when there is reasoning that does not correspond to the image.

\subsection{Joint Cross-Lingual and Cross-Modal Hallucinations}

Compared to cross-lingual and cross-modal hallucinations, joint Cross-Lingual and Cross-Modal Hallucinations involve generating responses in multiple languages that are inconsistent with both the visual content and each other. As shown in Figure~\ref{intro} (c), given an input image $I$, a question $Q$, and a prompt $P_{t}$ to produce answers in target languages, the MLLMs aim to generate responses $\mathcal{A}_{t}$ that accurately represent $I$ in response to $Q$ and are semantically consistent with their meanings:

\begin{equation}
	\mathcal{A}_{t} = \underset{\mathcal{A}_{t}'}{\operatorname{argmax}} \ P(\mathcal{A}_{t}'\ |\ Q, I, P_{t}),
\end{equation}
where $P(\mathcal{A}_{t}'\ |\ Q, I, P_{t})$ represents the joint probability of generating the answers given the inputs $Q$, $I$, and the prompt $P_{t}$.
Joint Cross-Lingual and Cross-Modal Hallucinations occur when the model does not respond in the target language as instructed, and the response includes reasoning that fails to reflect the content of the image.

\section{The \texttt{CCHall} Benchmark}

This section provides a detailed overview of the construction process of the \texttt{CCHall} benchmark, as illustrated in Figure~\ref{dataset}. The process primarily consists of the following four key components:

\begin{figure*}[t]
	\centering
	\includegraphics[width=160mm]{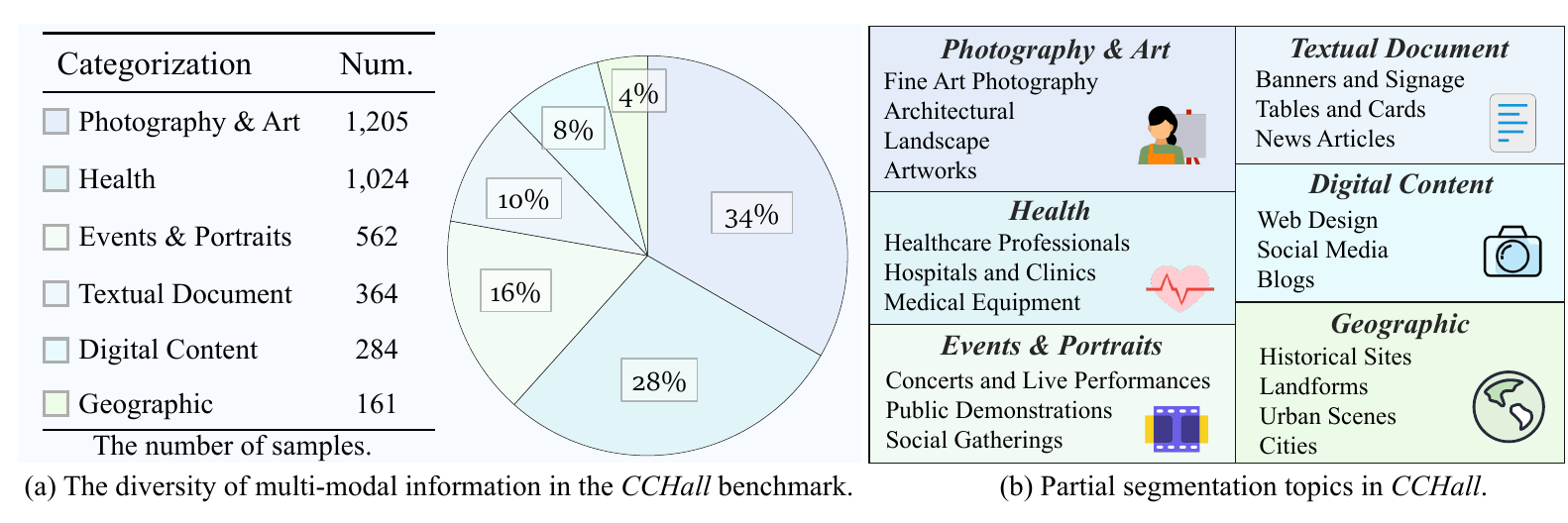}
	\vspace{-1\baselineskip}
	\caption{Presentation of data in \texttt{CCHall}: (a) The diversity of multi-modal data as represented by CLIP-based \cite{radford2021learning} classification. (b) Display of part of the detailed topics in \texttt{CCHall}.}
	\vspace{-0.5\baselineskip}
	\label{image_diversity}
\end{figure*}

\subsection{Raw Multi-modal Dataset Selection}
\label{raw}
To evaluate the hallucination behaviors of MLLMs from multiple perspectives, we have integrated two widely used multimodal tasks, Visual Question Answering (VQA) and Image Captioning (IC) \cite{rohrbach2018object,li2023evaluating, gunjal2024detecting}, as illustrated in Figure~\ref{dataset}~(a).

\noindent
\textit{\textbf{Visual Question Answering Task}}: Our VQA task extends GQA \cite{hudson2019gqa} and AMBER \cite{wang2023llm} dataset to assess reasoning and spatial understanding. GQA enriches scene descriptions, while AMBER covers 337 object categories. These datasets pose challenging VQA questions requiring advanced reasoning.

\noindent
\textit{\textbf{Image Captioning Task}}: The IC task evaluates hallucination detection in image descriptions, using the XM3600 \cite{thapliyal2022crossmodal} and xFlickr\&Co \cite{bugliarello2022iglue} datasets. XM3600 features diverse global images, while xFlickr\&Co focuses on everyday scenes, ensuring cultural diversity and minimal bias.

\vspace{-0.1\baselineskip}
\subsection{Cross-modal Hallucination Data Construction}
\label{Cross-modal Hallucination Data}

As shown in Figure~\ref{dataset}~(b), the process involves three steps to construct cross-modal hallucination data:

\noindent
\textit{\textbf{Automatic Sample Removal}}: To prepare the data for hallucination generation, we first filter out abnormal data, including instances of mismatched image-text pairs, images without corresponding text, or text without corresponding images, ensuring the dataset is clean and consistent.

\noindent
\textit{\textbf{Data Selection}}:
We define cross-modal hallucination as significant discrepancies between the outputs of MLLMs and the actual presence of objects in input images. Therefore, we select questions from the GQA \cite{hudson2019gqa} and AMBER \cite{wang2023llm} datasets related to object existence to assess the model's accuracy in object reasoning. To minimize redundancy, we ensure that each object appears no more than twice. For the Image Captioning task, we utilize the XM3600 \cite{thapliyal2022crossmodal} and xFlickr\&Co~\cite{bugliarello2022iglue} datasets, with English as the reference language. Finally, we randomly sample 900 entries from each of the filtered datasets, yielding a total of 3600 entries, to construct the dataset for generating hallucinated data.

\noindent
\textit{\textbf{Hallucination Data Generation}}:
We use the Gemini-1.5-Pro \cite{team2024gemini} to generate hallucinated data by inserting semantically similar but incorrect entities into image captions, increasing the challenge of \texttt{CCHall}. Real questions and answers were prepared for each image, along with prompts for the model. Gemini-1.5-Pro compared the image with the real answers, embedding misleading nouns not present in the image to ensure natural-sounding hallucinations. For a detailed description, please refer to Appendix~\ref{sec:Generating Hallucinated Data}.

\subsection{Cross-lingual Hallucination Data Construction}
\label{Cross-lingual Hallucination Data}

As shown in Figure~\ref{dataset}~(c), the process for constructing cross-lingual hallucination data is as follows:

\noindent
\textit{\textbf{Data Translation}}: Following \citet{conneau2018xnli,heredia2024xnlieu,hasan2024nativqa}, we apply machine translation to the text data and then conduct manual verification of the data. Specifically, following \citet{guerreiro2023hallucinations}, we categorize the languages into low, medium, and high-resource groups based on the availability of resources and select the three languages with the lowest error rates from each group to maximize translation accuracy. The final selections are as follows: Croatian (\textsc{hr}), Welsh (\textsc{cy}), and Swahili (\textsc{sw}) as low-resource languages; Czech (\textsc{cs}), Dutch (\textsc{nl}), and Swedish (\textsc{sv}) as medium-resource languages; and French (\textsc{fr}), Spanish (\textsc{es}), and Portuguese (\textsc{pt}) as high-resource languages. Google Translate is then used to translate the content into these target languages, resulting in an intermediate dataset.

\noindent
\textit{\textbf{Automatic Combination}}:
After validating the intermediate dataset, we shuffle the data within each subset to eliminate ordering bias and ensure an even distribution of sample types. Next, we refine the selection process by using English as the anchor language, paired with a randomly selected target language for each image. One response is in English, and the other is in a different language.

\noindent
\textit{\textbf{Human Recheck}}:
After generating the hallucinated data and translations, we conduct a human review to ensure data quality and overall accuracy. We focus on two key aspects: (1)~\textit{\underline{Verification of Hallucinated Data}}: We check if the generated data meets the hallucination criteria. This involves verifying that the sentences follow the expected structure, length, and logic and contain the intended elements aligned with the task. (2)~\textit{\underline{Accuracy of Translated Data}}: We verify if the translated data accurately reflects the original meaning and context. This step ensures that translations remain faithful to the intended scenario, avoiding errors from language differences. The details of the check can be found in Appendix~\ref{sec:Human Recheck}.

\begin{table*}[t]
\setlength{\tabcolsep}{7pt}
\centering
\begin{adjustbox}{width=0.98\textwidth}
\begin{tabular}{lcccccccccc}
\toprule
\multirow{2}{*}{Model}  &   \multicolumn{2}{c}{\textbf{AMBER}} & \multicolumn{2}{c}{\textbf{GQA}} & \multicolumn{2}{c}{\textbf{xFlickr\&CO}} & \multicolumn{2}{c}{\textbf{XM3600}} & \multicolumn{2}{c}{\textbf{AVG}} \\
	\cmidrule{2-11}
           &  \textbf{Acc}  &  \textbf{Macro-F1}   & \textbf{Acc} & \textbf{Macro-F1}  &   \textbf{Acc} &  \textbf{Macro-F1}   &  \textbf{Acc} & \textbf{Macro-F1}  &  \textbf{Acc} & \textbf{Macro-F1}  \\
        \midrule
        \rowcolor[rgb]{ .980,  .980,  .980} 
        Random    & 25.1 & 30.0 & 25.0 & 29.4 & 24.9 & 30.3 & 25.1 & 29.9 & 25.0 & 29.9\\
		\midrule
        \rowcolor[rgb]{ .970, .978, .999 }
        \multicolumn{11}{c}{\rule{0pt}{2ex}\textit{\fontsize{11pt}{11pt}\selectfont InternVL2-8B} \cite{chen2024far}}     \\  
		\midrule
		\rowcolor[rgb]{ .970, .978, .999 }
		\texttt{Direct} \cite{chen2024far}  & 29.1 & 38.1  & 29.9 & 38.6  & 38.3 & 47.6  & 38.8 & 47.4  & 34.0 & 42.9\\
		\rowcolor[rgb]{ .970, .978, .999 } 
		\texttt{CoT} \cite{kojima2022large}  & 31.3 & \underline{40.0} & \underline{33.6} & \underline{42.1}  & \underline{41.6} & 48.0  & \underline{40.1} & 47.6  & \underline{36.7} & \underline{44.4}\\
		\rowcolor[rgb]{ .970, .978, .999 }
		\texttt{SRO} \cite{lin2024interpreting}   & 30.3 & \underline{40.0} & 31.1 & 40.7  & 41.2 & \underline{48.6}  & 37.7 & 47.1  & 35.1 & 44.1\\
		\rowcolor[rgb]{ .970, .978, .999 }
		\texttt{VDGD} \cite{ghosh2024vdgd}  & \underline{33.2} & \underline{40.0} & 30.2 & 37.8  & 36.7 & 44.5  & 37.4 & 44.7  & 34.4 & 41.7\\
		\rowcolor[rgb]{ .970, .978, .999 }
		\texttt{HalluciMAD}  \cite{lin2024interpreting} & 29.9 & 38.6 & 30.0 & 39.0  & 37.9 & 45.9  & 39.6 & \underline{47.7}  & 34.3 & 42.8\\
		\midrule
        \rowcolor[rgb]{ .970, .978, .999 }
        \multicolumn{11}{c}{\rule{0pt}{2ex}\textit{\fontsize{11pt}{11pt}\selectfont Llama-3.2-11B-Vision-Instruct} \cite{dubey2024llama}} \\
		\midrule
		\rowcolor[rgb]{ .970, .978, .999 }
		\texttt{Direct} \cite{dubey2024llama}  & 31.6  & 38.8 & 32.1 & 38.9  & 35.4 & 43.2  & 43.3 & 49.4 & 35.6 & 42.6\\
		\rowcolor[rgb]{ .970, .978, .999 }
		\texttt{CoT} \cite{kojima2022large}  & 32.0  & 40.6 & 34.3 & 40.9  & 43.6 & 51.6  & \underline{46.4} & \underline{54.0} & 39.1 & \underline{46.8}\\
		\rowcolor[rgb]{ .970, .978, .999 }
		\texttt{SRO} \cite{lin2024interpreting}  & 32.1  & 40.6 & 34.4 & 40.9  & \underline{43.7} & \underline{51.7} & \underline{46.4} & \underline{54.0} & \underline{39.2} & \underline{46.8}\\
		\rowcolor[rgb]{ .970, .978, .999 }
		\texttt{VDGD} \cite{ghosh2024vdgd}  & \underline{34.0} & \underline{41.2} & \underline{35.6} & \underline{42.1}  & 36.6 & 45.4 & 42.4 & 50.7 & 37.1 & 44.9\\
		\rowcolor[rgb]{ .970, .978, .999 }
		\texttt{HalluciMAD}  \cite{lin2024interpreting} & 29.7 & 38.7 & 32.6 & 40.3  & 36.4 & 45.1  & 34.8 & 42.8 & 33.4 & 41.7\\
		\midrule
        \rowcolor[rgb]{ .970, .978, .999 }
        \multicolumn{11}{c}{\rule{0pt}{2ex}\textit{\fontsize{11pt}{11pt}\selectfont Qwen2-VL-7B-Instruct} \cite{wang2024qwen2}} \\
		\midrule
		\rowcolor[rgb]{ .970, .978, .999 }
		\texttt{Direct} \cite{wang2024qwen2}  & 36.2 & 39.5 & 33.3 & 38.3  & 42.9 & 46.7  & 39.9 & 44.4  & 38.1 & 42.2\\
		\rowcolor[rgb]{ .970, .978, .999 }
		\texttt{CoT} \cite{kojima2022large}  & 38.6 & 43.4 & 33.9 & \underline{39.0}  & 48.3 & 52.1  & \underline{48.4} & \underline{52.5}  & 42.3 & \underline{46.7}\\
		\rowcolor[rgb]{ .970, .978, .999 }
		\texttt{SRO} \cite{lin2024interpreting}  & \underline{38.7} & \underline{43.5} & 34.1 & 34.8  & \underline{48.4} & \underline{52.3}  & \underline{48.4} & 52.4  & \underline{42.4} & 45.7\\
		\rowcolor[rgb]{ .970, .978, .999 }
		\texttt{VDGD} \cite{ghosh2024vdgd}  & 34.8 & 36.3 & \underline{36.3} & 38.8  & 41.1 & 44.2  & 40.0 & 44.9  & 38.1 & 41.1\\
		\rowcolor[rgb]{ .970, .978, .999 }
		\texttt{HalluciMAD}  \cite{lin2024interpreting} & 37.0 & 42.5 & 31.4 & 38.1  & 38.0 & 45.4  & 38.6 & 44.7  & 36.3 & 42.7\\
		\midrule
        \rowcolor[rgb]{ .970, .978, .999 }
        \multicolumn{11}{c}{\rule{0pt}{2ex}\textit{\fontsize{11pt}{11pt}\selectfont Pixtral-12B-2409} \cite{agrawal2024pixtral}} \\
		\midrule
		\rowcolor[rgb]{ .970, .978, .999 }
		\texttt{Direct} \cite{agrawal2024pixtral}  & 23.9 & 38.0 & 34.8 & 45.5  & 43.3 & 51.3  & 38.0 & 48.9  & 35.0 & 45.9\\
		\rowcolor[rgb]{ .970, .978, .999 }
		\texttt{CoT} \cite{kojima2022large}  & 40.7 & 46.0 & 42.7 & 47.9  & 48.7 & 54.2  & 53.3 & 59.0  & 46.3 & 51.8\\
		\rowcolor[rgb]{ .970, .978, .999 }
		\texttt{SRO} \cite{lin2024interpreting}  & 43.0 & 47.8 & 41.3 & 48.1  & 50.6 & 55.5  & 51.6 & 57.8  & 46.6 & 52.3\\
		\rowcolor[rgb]{ .970, .978, .999 }
		\texttt{VDGD} \cite{ghosh2024vdgd}  & 43.4 & 43.1 & 45.1 & 47.3  & 47.3 & 48.4  & 56.4 & 58.2  & 48.1 & 49.3\\
		\rowcolor[rgb]{ .970, .978, .999 }
		\texttt{HalluciMAD}  \cite{lin2024interpreting} & \underline{46.3}  & \underline{49.8} & \underline{45.2}  & \underline{51.4}  & \underline{57.1}  & \underline{61.4}  & \underline{58.7}  & \underline{63.2}  & \underline{51.8} & \underline{56.4} \\
		\midrule
        \rowcolor[rgb]{ .961, .985, .970 }
        \multicolumn{11}{c}{\rule{0pt}{2ex}\textit{\fontsize{11pt}{11pt}\selectfont Gemini-1.5-Flash}  \cite{team2024gemini}} \\
		\midrule
		\rowcolor[rgb]{ .961, .985, .970 }
		\texttt{Direct} \cite{team2024gemini}  & 41.7  & 44.7 & 37.0 & 39.2 & 49.2 & 50.5 & 50.1 & 51.9 & 44.5 & 46.6 \\
		\rowcolor[rgb]{ .961, .985, .970 }
		\texttt{CoT} \cite{kojima2022large}  & 49.2 & 52.3 & 53.2 & 54.7 & 56.4 & 58.0 & 60.6 & 62.2 & 54.9 & 56.8 \\
		\rowcolor[rgb]{ .961, .985, .970 } 
		\texttt{SRO} \cite{lin2024interpreting}   & 49.0  & 52.9 & 51.6 & 53.4 & 55.7 & 58.3 & 58.2 & 60.8 & 53.6 & 56.4 \\
		\rowcolor[rgb]{ .961, .985, .970 }
		\texttt{VDGD} \cite{ghosh2024vdgd}  & \underline{58.9}  & \underline{60.3} & 50.6 & 52.8 & 52.7 & 54.0 & 60.6 & 62.4 & 55.7 & 57.4 \\
		\rowcolor[rgb]{ .961, .985, .970 }
		\texttt{HalluciMAD}  \cite{lin2024interpreting} & 52.2  & 54.5 & \underline{59.0} & \underline{60.6} & \underline{61.6}  & \underline{63.9} & \underline{63.7}  & \underline{64.9} & \underline{59.1} & \underline{61.0} \\
		\midrule
        \rowcolor[rgb]{ .961, .985, .970 }
        \multicolumn{11}{c}{\rule{0pt}{2ex}\textit{\fontsize{11pt}{11pt}\selectfont GPT-4o} \cite{achiam2023gpt}}  \\
		\midrule
		\rowcolor[rgb]{ .961, .985, .970 }
		\texttt{Direct} \cite{achiam2023gpt}  & 57.1  & 63.2 & 56.1  & 62.1 & 72.2  & 76.4 & 82.9  & 84.3 & 67.1  & 71.5 \\
		\rowcolor[rgb]{ .961, .985, .970 }
		\texttt{CoT} \cite{kojima2022large}  & 68.1  & 70.0 & 66.9  & 69.2 & 81.1  & 83.4 & 83.1  & 84.8 & 74.8  & 76.8 \\
		\rowcolor[rgb]{ .961, .985, .970 }
		\texttt{SRO} \cite{lin2024interpreting}  & 63.2  & 65.0 & 57.2  & 62.7 & 76.7  & 79.3 & 84.7  & 86.4 & 70.4  & 73.4 \\
		\rowcolor[rgb]{ .961, .985, .970 }
		\texttt{VDGD} \cite{ghosh2024vdgd}  & 64.7  & 66.6 & 56.3  & 62.5 & 72.3  & 77.0 & 83.2  & 85.6 & 69.1  & 73.0 \\
		\rowcolor[rgb]{ .961, .985, .970 }
		\texttt{HalluciMAD}  \cite{lin2024interpreting} & \textcolor{gold}{\ding{202}}  \textbf{70.9}  & \textcolor{gold}{\ding{202}}  \textbf{71.9}  & \textcolor{gold}{\ding{202}} \textbf{68.6}   & \textcolor{gold}{\ding{202}} \textbf{70.3}  & \textcolor{gold}{\ding{202}} \textbf{84.1}   & \textcolor{gold}{\ding{202}} \textbf{85.6}  & \textcolor{gold}{\ding{202}} \textbf{86.4}   & \textcolor{gold}{\ding{202}} \textbf{87.3}  & \textcolor{gold}{\ding{202}} \textbf{77.5}   & \textcolor{gold}{\ding{202}} \textbf{78.8}  \\
		\bottomrule
	\end{tabular}
	\end{adjustbox}
    \caption{The experimental results of Acc. (\%) and Macro-F1 score on MLLMs. The ``Random'' refers to the average performance obtained from three separate random selections. \raisebox{0.7mm}{\colorbox{myblue}{\textcolor{myblue}{\rule{0.7mm}{0.7mm}}}} represents the performance of open-source MLLMs, and \raisebox{0.7mm}{\colorbox{mygreen}{\textcolor{mygreen}{\rule{0.7mm}{0.7mm}}}} represents the performance of closed-source MLLMs. The \underline{underline} indicates better performance in the MLLM. \textcolor{gold}{\ding{202}}  represents the \textbf{Best} performance. The complete results are shown in Table~\ref{main results all}.}
    \vspace{-0.8\baselineskip}
    \label{main results}
\end{table*}

\subsection{Cross-modal and Cross-lingual Hallucination Dataset}
\label{CCHall Dataset}
The \underline{C}ross-modal and \underline{C}ross-lingual \underline{Hall}ucination Dataset (\texttt{CChall}) is shown in Figure~\ref{dataset}~(d). Please refer to the Appendix~\ref{sec:Error Analysis} for specific examples. Specifically, we have retained the original dataset names as the categories. In total, the categories are divided into four types: \textit{AMBER, GQA, xFlickr\&Co, and XM3600}. Additionally, we define four types of combinations based on hallucinations:
 
 \vspace{-1mm}

\begin{mybox}

\begin{itemize}
\item \textit{\textbf{Non-hallucination}}: Both the English responses and the answers in the other language are correct, with all answers matching the contents in the images.

\vspace{-2mm}

\item \textit{\textbf{Cross-lingual, non-cross-modal hallucination}}: The English answer is correct, but the non-English answer contains hallucinated objects not present in the images.

 \vspace{-2mm}
 
\item \textit{\textbf{Cross-modal, non-cross-lingual hallucination}}: All answers are hallucinatory, containing objects that do not appear in the images. Meanings in each language are identical.

 \vspace{-2mm}
 
\item \textit{\textbf{Cross-modal, cross-lingual hallucination}}: All the answers are hallucinated, containing various objects that are either absent from the images or incorrectly described. Moreover, the meanings conveyed in the non-English answers differ significantly from those in the English answers.

 \vspace{-1mm}
 
\end{itemize}

\end{mybox}

\noindent
A more detailed presentation of the \texttt{CCHall} attributes can be found in Appendix~\ref{sec:Four Hallucination Types}.

\begin{figure}[t]
	\centering
	\includegraphics[width=0.5\textwidth]{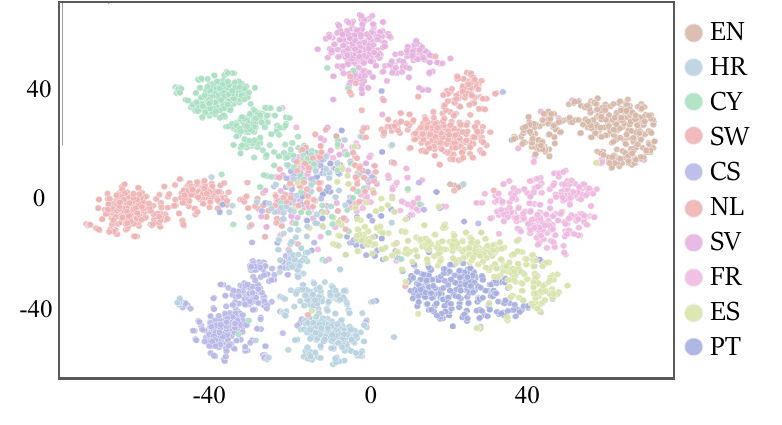}
	\vspace{-5mm}
	\caption{Visualization of the semantic feature coverage of all languages in \texttt{CCHall}, demonstrating the distribution and range of linguistic representations.
}
\vspace{-4mm}
	\label{type}
\end{figure}

\section{\texttt{CCHall} Analysis}
This section presents an analysis of \texttt{CChall}.
\noindent
\textbf{Multi-modal Diversity}: 
Following \citet{chen2024m}, we employ CLIP \cite{radford2021learning} to classify the images in the \texttt{CCHall}. The results, as shown in Figure~\ref{image_diversity}, reveal that the images in \texttt{CCHall} exhibit a wide variety of types. This diversity enables a more comprehensive coverage of the hallucination scenarios demonstrated by LLMs.

\noindent
\textbf{Multi-lingual Diversity}: To demonstrate the diversity of multilingualism, we use the CLIP~\cite{radford2021learning} text encoder to encode data in all languages and applied t-SNE for dimensionality reduction. As shown in Figure~\ref{type}, the language features cover nearly the entire range of semantic feature representations. This indicates that \texttt{CCHall} exhibits broad linguistic richness, enabling comprehensive detection of multilingual hallucinations.

\section{Experiments}
\subsection{Experiments Setting}
We evaluate several MLLMs on the \texttt{CCHall}, including \textit{GPT-4o} \cite{achiam2023gpt}, \textit{Gemini-1.5-Flash}~\cite{team2024gemini}, \textit{Llama-3.2-11B-Vision-Instruct} \cite{meta2024llama}, \textit{Qwen2-VL-7B-Instruct} \cite{wang2024qwen2}, \textit{Pixtral-12B-2409} \cite{agrawal2024pixtral}, and \textit{InternVL2-8B} \cite{chen2024far}. 
In addition to using \texttt{Direct} queries with MLLMs, we explore several strategies to mitigate hallucinations. Specifically, we implement \texttt{CoT} \cite{kojima2022large} to promote step-by-step reasoning, \texttt{SRO} \cite{lin2024interpreting} for self-reflection, and \texttt{VDGD} \cite{ghosh2024vdgd} to embed detailed image descriptions and better align reasoning. We apply \texttt{HalluciMAD} \cite{lin2024interpreting} to reduce hallucinations via multi-agent debate. All top-p and temperature parameters retain the default of MLLMs values within the specified range of [0, 1].

\vspace{-2mm}
\subsection{Results for \texttt{CCHall}}

Results are summarized in the Table~\ref{main results}. We have made the following observations:

\vspace{1mm}
\noindent
\textbf{\texttt{CCHall} is a challenging hallucination detection benchmark}: Our evaluation results show that the weakest-performing model is \textit{InternVL2-8B}, achieving an accuracy of 34.0\% using the \texttt{Direct} method. In contrast, the best-performing method, \texttt{HalluciMAD}, achieves an accuracy of 77.5\% using \textit{GPT-4o}. This suggests that \texttt{CCHall} is a highly challenging benchmark and that MLLMs still have significant room for further improvement in cross-lingual and cross-modal tasks.

\vspace{1mm}
\noindent
\textbf{The performance of MLLMs depends on their training strategy and parameter size}: The closed-source \textit{GPT-4o} and \textit{Gemini-1.5-Flash} exhibit superior performance, surpassing several open-source models. Additionally, performance variations exist among open-source models. Notably, the \textit{Qwen-2-VL-7B} outperforms both \textit{Llama-3.2-11B} and \textit{InternVL2-8B}. This highlights the vital role of training strategies in enhancing performance.

\vspace{1mm}
\noindent
\textbf{Strategies for mitigating hallucinations work in specific contexts}: Hallucination mitigation methods generally outperform the \texttt{Direct} approach, although their effectiveness can vary depending on the specific context. Basic methods, such as \texttt{CoT} and \texttt{SRO}, perform better in less powerful models, including \textit{InternVL2-8B}, \textit{Llama-3.2-11B-Vision}, and \textit{Qwen-2-VL-7B}. In contrast, more advanced methods, such as \texttt{VDGD} and \texttt{HalluciMAD}, are more effective in powerful models, including \textit{Pixtral-12B-2409}, \textit{Gemini-1.5-Flash}, and \textit{GPT-4o}.

\begin{figure*}[t]
	\centering
	\includegraphics[width=1.01\textwidth]{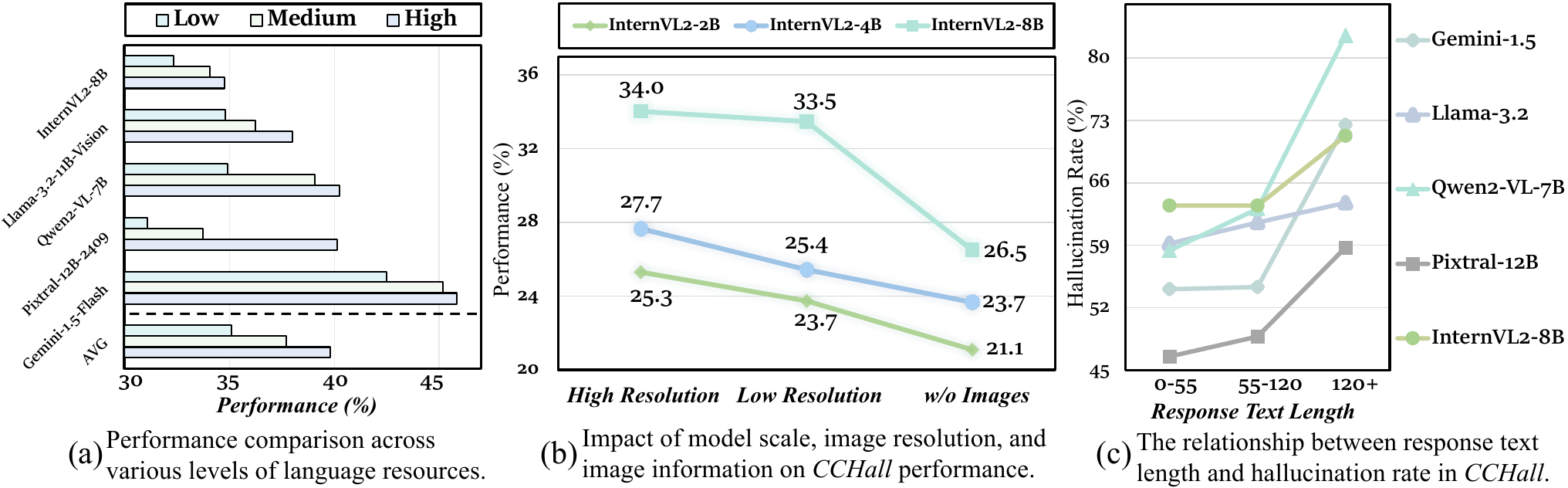}
	\caption{Analysis of the underlying causes of cross-lingual and cross-modal hallucinations in MLLMs.
	}
	\label{fig:a3}
\end{figure*}

\subsection{Analysis}
We conduct experiments to analyze how image quality, language resources, response length, and object-specific biases influence hallucination behavior and detection in \texttt{CCHall}.

\subsubsection{Models Perform Better in High-resource Languages than in Low-resource Ones}

To examine performance variations across languages, we follow \citet{guerreiro2023hallucinations} and categorize them into three groups: high-resource languages (\textsc{fr}, \textsc{es}, \textsc{pt}), medium-resource languages (\textsc{cs}, \textsc{nl}, \textsc{sv}), and low-resource languages (\textsc{hr}, \textsc{cy}, \textsc{sw}). As illustrated in Figure~\ref{fig:a3} (a), we observe that hallucination detection accuracy is highest for high-resource languages and lowest for low-resource languages. This disparity in performance is likely attributed to the limited availability of training data for low-resource languages, which significantly hampers the model’s ability to understand contextual nuances, leading to increased difficulty in accurate hallucination detection.

\subsubsection{High-resolution Images and Larger Models Enhance Task Performance} 

To explore the impact of image quality and model parameters on hallucination performance, we analyze the effects of high-resolution, low-resolution, and no-image (w/o Images) inputs on model performance. At the same time, we use InternVL of different sizes to examine the effect of model scale on hallucination. As shown in Figure~\ref{fig:a3} (b), performance declines as image resolution decreases, with a sharp drop when no image is provided, highlighting the importance of visual information in \texttt{CCHall} for reducing hallucinations. Furthermore, model performance declines as the number of parameters decreases (Performance: 8B > 4B > 2B).

\subsubsection{Longer Responses Generally Always Lead to Higher Hallucination Rates} 

We investigate the effect of response length on model performance. As shown in Figure~\ref{fig:a3} (c), there is a strong positive relationship between how long a response is and how often hallucinations occur. Notably, a substantial surge in hallucination rate is observed when the length of the response surpasses 120 words, indicating a critical threshold beyond which the model's output becomes significantly less reliable. The objects that are susceptible to hallucination are detailed in Appendix~\ref{appendix obj}. This is likely because the model generates additional reasoning steps when the answer is uncertain. In the future, incorporating reflective steps into the model may help mitigate hallucinations.

\vspace{-1mm}

\subsection{Exploration}
This section explores two factors for improving multimodal performance: multi-language prompts and external tools for detecting hallucinations.

\subsubsection{Multilingual Prompt Exploration}
\begin{figure}[t]
	\centering
	\includegraphics[width=0.49\textwidth]{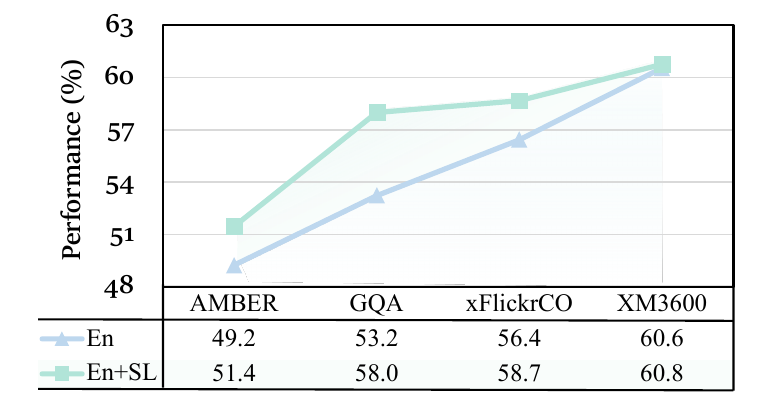}
	\caption{A comparison between using only English~(\textit{En}) and using English with an additional Source Language (\textit{EN+SL}) as context in \textit{Gemini-1.5-Flash}.
	}
	\label{multi_lang}
	\vspace{-2mm}
\end{figure}

We investigate the performance enhancement of the MLLMs in cross-lingual and cross-modal scenarios under bilingual context prompts. The results are shown in Figure~\ref{multi_lang}, incorporating the Source Language and English (\textit{En+SL}) consistently improves hallucination detection accuracy across datasets, compared to using only English~(\textit{En}). Specifically, in the two VQA tasks, accuracy increases by 2.2\% in the AMBER dataset and by 4.8\% in GQA. This improvement suggests that bilingual prompts provide better linguistic context and help effectively mitigate biases associated with relying solely on English. Future research should focus on further optimizing bilingual prompts and evaluating their effectiveness across additional language pairs.

\subsubsection{Framework Adaptation Exploration}

\begin{figure}[t]
	\centering
	\includegraphics[width=0.49\textwidth]{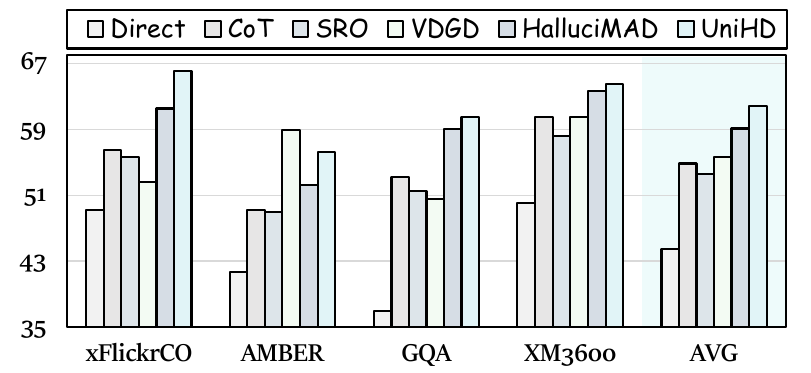}
	\caption{The \texttt{UniHD}, which utilizes external tools, is compared with other methods in \textit{Gemini-1.5-Flash}.
}
	\label{unihd}
\end{figure}

To investigate how external tools and web search improve the mitigation of hallucinations, we integrate the \texttt{UniHD} \cite{chen2024unified} framework. As shown in Figure~\ref{unihd}, adapting the \texttt{UniHD} framework for \texttt{CCHall} allows the model to achieve the highest accuracy across all datasets, surpassing previous methods such as \texttt{VDGD} and \texttt{HalluciMAD}. Across all datasets, it shows an average improvement of 2.7\% over \texttt{HalluciMAD}. By validating visual claims using external tools and web resources, \texttt{UniHD} bridges the gap between model predictions and real-world data, resulting in significant improvements over methods that rely solely on the internal knowledge reasoning of MLLMs. The experimental details can be found in Appendix~\ref{sec:Framework Adaptation Exploration}.

\section{Related Work}
LLMs have recently made rapid progress~\cite{team2024gemini,qin2025survey,chen2025reasoningerasurveylong}, achieving outstanding performance~\cite{zhang2024wrong}. However, LLMs often exhibit hallucinations, producing outputs that fail to align with the given inputs, particularly in cross-lingual or cross-modal settings \cite{benkirane2024machine,favero2024multi}. This has prompted several studies exploring hallucinations \cite{sriramanan2025llm}.

\noindent
\textbf{Cross-lingual Hallucinations Benchmark}: \citet{qiu2023detecting} introduce mFACT to assess the faithfulness of non-English summaries, revealing that LLMs are more prone to hallucination in languages other than English. \citet{dale2023halomi} release an annotated dataset covering hallucinations and omissions across 18 translation directions, with varying levels of hallucination severity. \citet{herrlein2024anhalten} extend English hallucination detection to German and apply it to long-context scenarios. \citet{son2024mmevalmultilingualmetaevaluationbenchmark} introduce MM-Eval, a multilingual benchmark covering 18 languages, to evaluate LLMs as evaluators. \citet{li2024xtrustmultilingualtrustworthinesslarge} introduce XTRUST, the multilingual trustworthiness benchmark, covering 10 languages and various topics like hallucination, misinformation, and fairness.

\noindent
\textbf{Cross-modal Hallucinations Benchmark}: \citet{liu2023hallusionbench} introduce a benchmark for image-context reasoning using image-question pairs created by experts. \citet{leng2024curse} propose the ``Curse of Multi-Modalities'' benchmark to evaluate multimodal hallucinations in MLLMs. \citet{yan2024evaluating} propose a framework for evaluating the reliability and validity of hallucination benchmarks. \citet{gunjal2024detecting} introduce M-HalDetect, a multimodal dataset with 16k VQA annotations for hallucination detection in LVLMs. \citet{guan2024hallusionbench} introduce HallusionBench, a benchmark for image-context reasoning in LVLMs with 346 images and 1129 questions. \citet{chen2024unified} introduce a multi-modal meta-evaluation benchmark for assessing hallucination detection in MLLMs.

Compared to previous work, we introduce the \textbf{C}ross-lingual and \textbf{C}ross-modal \textbf{Hall}ucinations benchmark (\texttt{CCHall}). To the best of our knowledge, this is the first effort to explore joint cross-lingual and cross-modal hallucinations.

\section{Conclusion}

We introduce a benchmark, \texttt{CCHall}, for the detection of joint cross-lingual and cross-modal hallucinations. We analyze a wide range of MLLMs and various hallucination mitigation strategies. Our findings demonstrate that current strategies are still insufficient in effectively overcoming cross-lingual and cross-modal hallucinations. Specifically, they often struggle to effectively handle the complexities that arise when both cross-lingual and cross-modal factors interact. We hope \texttt{CCHall} can serve as valuable data for evaluating LLMs in joint cross-lingual and cross-modal scenarios.

\section*{Limitations}

We introduce \texttt{CCHall}, a cross-lingual and cross-modal hallucination detection benchmark, and perform comprehensive experiments across various MLLMs. Current MLLMs are mainly confined to two modalities: text and image, with only a few extending to the Audio/Speech modality \cite{borsos2023audiolm, kuan2024understanding}. Consequently, our work concentrates on the text and image modalities. We hope that future MLLMs will incorporate a broader range of modalities, bringing us one step closer to achieving Artificial General Intelligence (AGI). Additionally, since the hallucinated data in our dataset is generated by MLLMs and the multilingual data is obtained through translation, some errors may persist despite manual verification.

\section*{Acknowledgments}
This work was supported by the National Natural Science Foundation of China (NSFC) via grant 62306342 and 62376130. This work was sponsored by the Excellent Young Scientists Fund in Hunan Province (2024JJ4070), the Science and Technology Innovation Program of Hunan Province under Grant 2024RC3024 and CCF-Zhipu Large Model Innovation Fund (NO.CCFZhipu202406). This work was supported by Key Laboratory of Data Intelligence and Advanced Computing in Provincial Universities, Soochow University. This work was supported by the Key Laboratory of Computing Power Network and Information Security, Affiliated with Ministry of Education, Qilu University of Technology (Shandong Academy of Sciences) (No.2023ZD032).
We are grateful for resources from the High Performance Computing Center of Central South University.

\bibliography{custom}

\newpage

\appendix
\section*{Appendix}
\section{\texttt{CCHall} Construction Details}
\label{sec:appendix}

\subsection{Generating Hallucinated Data}
\label{sec:Generating Hallucinated Data}

\begin{figure*}[t]
	\centering
	\includegraphics[width=\textwidth]{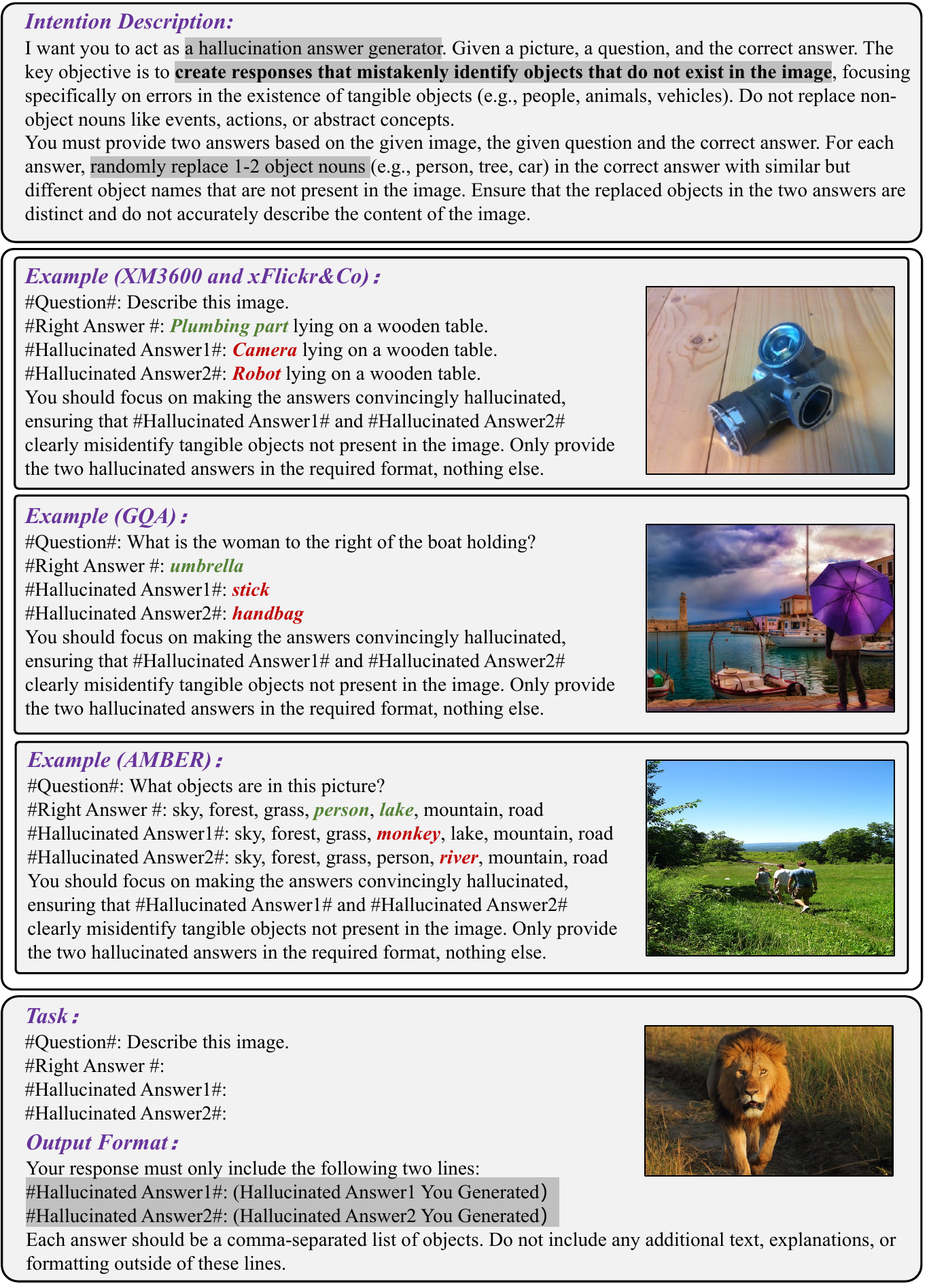}
	\caption{
    Structure of the Prompt for Generating Hallucinated Sentences}
	\label{fig:generate_halu_prompt}
\end{figure*}

Before generating data, we first prepare corresponding real questions and answers for each image and meticulously design specific prompts to guide Gemini-1.5-Pro in generating hallucinated sentences. As shown in Figure~\ref{fig:generate_halu_prompt}, the prompt includes the following elements:

\begin{mybox}
\
\begin{enumerate}
  \item \textit{\textbf{Image}}: The inclusion of the image ensures that the model effectively recognizes and processes visual content, thereby facilitating more accurate multi-modal alignment.

\item \textit{\textbf{Question and Real Answer}}:
These provide the model with an accurate semantic context, enabling it to understand the requirements for generating hallucinated data.

\item \textit{\textbf{Examples}}:
By providing examples, the prompt clarifies the generation target for the model, ensuring that the hallucinated sentences maintain logical consistency and follow the intended structure and content. As shown in Figure~\ref{fig:generate_halu_prompt}, these are the provided examples used to construct A and B.
\end{enumerate}
\end{mybox}

Incorporating these elements helps the model better comprehend the input context, ensuring that the generated hallucinated sentences appropriately combine information from the image, question, and real answer to produce the expected output. This design prevents the hallucinations from deviating from the actual context, ensuring high quality in the generated content.

During the generation process, we require the hallucinated sentences to closely match the real answers in both length and structure. Specifically, the Gemini-1.5-Pro, when processing images, is instructed to compare the real entities in the answers with the content in the images. It then integrates entities that do not exist in the image into the generated hallucinated sentence.

This approach ensures that the hallucinated data maintains a fundamental level of naturalness and consistency with the real answer. If the structure of the hallucinated sentence deviates significantly from the real answer, it becomes more easily identifiable as an anomaly by both models and human users. Thus, maintaining similarity in structure and length makes the hallucinated sentence appear more natural and harder to detect. Moreover, this strategy ensures that the hallucinated data we generate is highly deceptive. Structurally similar hallucinated sentences can mask the inserted false entities, ensuring that the hallucinated content maintains fluency similar to the real answer, thus increasing the difficulty for models during evaluation.

Overall, this process ensures that the hallucinated sentences semantically approximate the real answers, while preserving structural consistency, thus enhancing their deceptiveness and improving evaluation effectiveness.

\subsection{Cross-lingual Hallucination Data Construction}
\label{sec:Human Recheck}
\textbf{Human Recheck:}
\noindent
After generating the hallucinated data and performing translations, we carefully conduct a human recheck to ensure the quality and accuracy of the data. Specifically, we consistently employ the back-translation method \cite{miyabe2015evaluation, lee2021examining} to thoroughly review all the data, focusing on two key aspects:

\begin{mybox}
\
\begin{enumerate}
  \item \textit{\textbf{Verification of Hallucination Data}}: We check whether the generated hallucinated data meets our requirements. This involves ensuring that the hallucinated sentences adhere to the expected structure, length, and logic, while also verifying that they contain the intended hallucinated elements in alignment with the task.
  \item \textit{\textbf{Accuracy of Translated Data}}: We also check whether the translated data accurately reflects the original meaning and context. This step ensures that the translations remain faithful to the intended hallucination scenario, avoiding any misinterpretations or errors that arise from language differences.
\end{enumerate}
\end{mybox}

\noindent
In terms of details, we provide the reviewers with a specific scoring rubric as a clear guide for evaluation. The scoring criteria are as follows:

\begin{enumerate}
  \item [•] 0-60 points: Only a small portion of the original meaning is conveyed. The word choice is inaccurate, and there are a significant number of severe language errors.
  \item [•] 60-70 points: The meaning of the original text is conveyed. Word choice is somewhat inaccurate, and there are quite a few language errors, some of which are serious.
  \item [•] 70-80 points: The meaning of the original text is generally conveyed. The text is smooth and coherent, with no major language errors.
  \item [•] 80-90 points: The text conveys the original meaning clearly, with minor errors that don’t hinder comprehension. The writing is smooth and coherent, though some word choices may be slightly imprecise or awkward.
  \item [•] 90-100 points: The original meaning is accurately conveyed. The word choice is appropriate, the writing flows smoothly, and there are virtually no language errors.
\end{enumerate}

\noindent
Each data is reviewed by three evaluators, and the final score for the data is obtained by averaging the scores from all three reviewers. If a data point receives a score below 80, we regenerate and retranslate it. The data will then be re-evaluated and re-scored until it passes the review. A total of only 11 data points received a score below 80, and we have already made corrections to them.

In our review, the average score across all data points is 87.1. The manual check guarantees the accuracy, consistency, and quality of the generated data and translations, significantly minimizing the risk of errors, inconsistencies, or biases that could affect the model's overall performance.

\subsection{Four Hallucination Types}
\label{sec:Four Hallucination Types}

Based on the definitions of four hallucination types, we carefully select and combine the intermediate dataset after randomization and balancing. Specifically, we use English as the anchor language and select two answers for each image in different languages. For each image, one English answer is randomly chosen, along with another in a different language. The selected English answer can be either the correct answer or a hallucinated one generated by a previous model. According to the hallucination definitions, we have the following four combinations:

\begin{mybox}

\begin{enumerate}
  \item [1.] \textit{\textbf{Non-hallucination}}: If both the English answer and the answer in another language are correct, meaning that the objects mentioned in the sentence match those in the image and are consistent with it, the answer is classified as non-hallucination.
\end{enumerate}
\end{mybox}

\begin{mybox}

\begin{enumerate}
	\item [2.] \textit{\textbf{Cross-lingual, non-cross-modal hallucination}}: If the English answer is correct but the answer in another language is hallucinated — meaning the English answer is consistent with the image while the objects in the non-English answer do not appear in the image — this is classified as cross-lingual non-cross-modal hallucination.
  \item [3.] \textit{\textbf{Cross-modal, non-cross-lingual hallucination}}: If the English answer is hallucinated and the answer in another language is identical, meaning both answers contain the same hallucinated objects not present in the image, this is classified as cross-modal non-cross-lingual hallucination.
  \item [4.] \textit{\textbf{Cross-modal, cross-lingual hallucination}}: If the English answer is hallucinatory and the answer in another language is inconsistent with it, meaning that the two hallucinatory answers differ, with each containing different objects that are either not actually present in the image or are incorrectly described, this is classified as a cross-modal, cross-lingual hallucination.
\end{enumerate}
\end{mybox}

This combination process ensures that we generate a diverse set of hallucination types, while maintaining consistency in image-object relationships, thus providing a robust and high-quality dataset for training and evaluation.

\subsection{Prompt Construction for Experimentation}
\begin{figure*}[t]
	\centering
	\includegraphics[width=\textwidth]{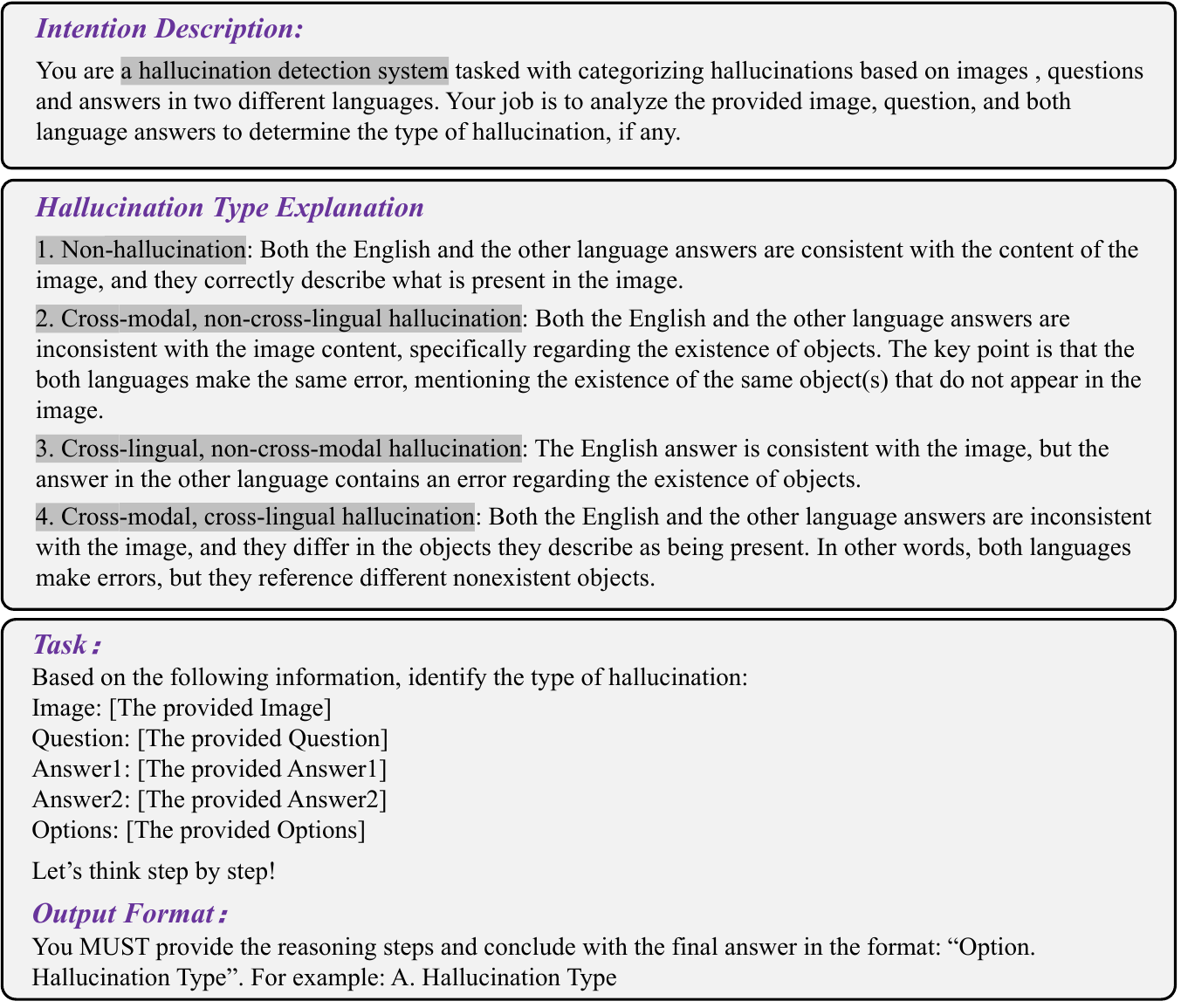}
	\caption{
    Example of the Prompt Used for Hallucination Detection and Classification}
	\label{fig:test_prompt}
\end{figure*}

During the experimental phase, we test different models on our \texttt{CCHall}. As shown in Figure~\ref{fig:test_prompt}, our prompt consists of four key components: Intention Description, Hallucination Type Explanation, Task Description, and Output Format.

\begin{figure*}[t]
	\centering
	\includegraphics[width=\textwidth]{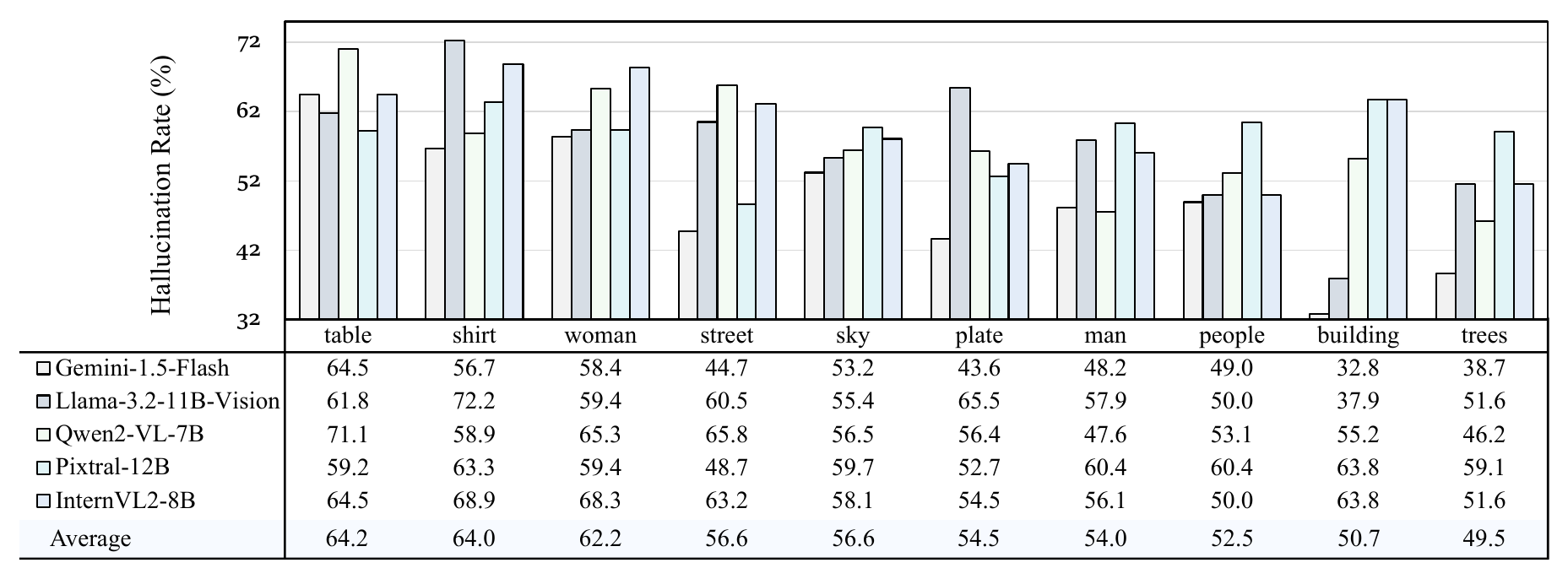}
	\caption{Analysis of the relationship between object presence and hallucination rate in the image captioning task.
	}
	\label{fig:obj}
\end{figure*}

\begin{enumerate}
  \item \textit{\textbf{Intention Description}}: This section establishes the model's role, clearly instructing it that its task is to detect hallucinations based on the given image, question, and answers in two different languages, and classify the hallucinations. The model selects the correct option from the provided choices. This helps set expectations for the model’s behavior, ensuring that it understands the specific goal of identifying and classifying hallucinations.

\item \textit{\textbf{Hallucination Type Explanation}}: This part defines four types of hallucinations: ``Non-hallucination'', ``Cross-lingual, non-cross-modal hallucination'', ``Cross-modal, non-cross-lingual hallucination'', and ``Cross-modal, cross-lingual hallucination''. Explaining the four types of hallucinations is critical because it provides the model with a clear framework to understand the specific nature of each hallucination type. By making these definitions explicit, the model can more effectively distinguish between different hallucination scenarios and improve its accuracy in detecting and classifying hallucinations. Without this explanation, the model struggles to identify subtle distinctions between hallucination types, which impacts its overall performance and classification accuracy.

\item \textit{\textbf{Task Description}}: This section provides concrete task details, including the image, question, two answers in different languages, and the available options. This supplies the model with all necessary input data, ensuring that it has everything needed to perform the task effectively and make informed decisions when selecting the correct answer.

\item \textit{\textbf{Output Format}}: The output format section specifies the required structure of the model’s response, designed for easy extraction using regular expressions. This standardization allows efficient evaluation of the model's performance, facilitates accurate calculations, and ensures consistent output formatting.
\end{enumerate}

This structured approach to prompt construction ensures that the model receives clear, unambiguous instructions at every stage of the task, improving consistency and reliability in hallucination detection and classification across different models.

\begin{table*}[t]
\setlength{\tabcolsep}{3pt}
\centering
\begin{adjustbox}{width=0.98\textwidth}
\begin{tabular}{lcccccccccc}
\toprule
\multirow{2}{*}{Model}  &   \multicolumn{2}{c}{\textbf{AMBER}} & \multicolumn{2}{c}{\textbf{GQA}} & \multicolumn{2}{c}{\textbf{xFlickr\&CO}} & \multicolumn{2}{c}{\textbf{XM3600}} & \multicolumn{2}{c}{\textbf{AVG}} \\
	\cmidrule{2-11}
           &  \textbf{Acc}  &  \textbf{Macro-F1}   & \textbf{Acc} & \textbf{Macro-F1}  &   \textbf{Acc} &  \textbf{Macro-F1}   &  \textbf{Acc} & \textbf{Macro-F1}  &  \textbf{Acc} & \textbf{Macro-F1}  \\
		\midrule
		\texttt{Direct} \cite{team2024gemini}  & 41.7  & 44.7 & 37.0 & 39.2 & 49.2 & 50.5 & 50.1 & 51.9 & 44.5 & 46.6 \\
		\texttt{CoT} \cite{kojima2022large}  & 49.2 & 52.3 & 53.2 & 54.7 & 56.4 & 58.0 & 60.6 & 62.2 & 54.9 & 56.8 \\ 
		\texttt{SRO} \cite{lin2024interpreting}   & 49.0  & 52.9 & 51.6 & 53.4 & 55.7 & 58.3 & 58.2 & 60.8 & 53.6 & 56.4 \\
		\texttt{VDGD} \cite{ghosh2024vdgd}  & \textbf{58.9}  & \textbf{60.3} & 50.6 & 52.8 & 52.7 & 54.0 & 60.6 & 62.4 & 55.7 & 57.4 \\
		\texttt{HalluciMAD}  \cite{lin2024interpreting} & 52.2  & 54.5 & 59.0 & \textbf{60.6} & 61.6  & \textbf{63.9} & 63.7  & 64.9 & 59.1 & 61.0 \\
		\midrule
		\rowcolor[rgb]{ .970, .978, .999 }
        \texttt{\textbf\texttt{\texttt{UniHD}}}  \cite{chen2024unified} & 56.2 & 52.4  & \textbf{60.6} & \textbf{60.6} & \textbf{66.1}  & 59.5 & \textbf{64.5}   & \textbf{65.7} &  \textbf{61.8} & \textbf{62.1} \\
 
            \bottomrule
            \end{tabular}
        \end{adjustbox}
      
        \caption{Comparison of the \texttt{UniHD} framework and other methods on Gemini. \textbf{Bold} represents the Best performance.}
        \label{unihd appendix}
\end{table*}

\section{Experiment Details}
\subsection{Main Result Details}
\label{sec:Main Result Details}
To conduct a more thorough and comprehensive evaluation of \texttt{CCHall}, we assess its performance on the following extensive models:  \textit{GPT-4o}~\cite{achiam2023gpt}, \textit{Gemini-1.5-Flash}~\cite{team2024gemini}, \textit{Llama-3.2-11B-Vision-Instruct} \cite{meta2024llama}, \textit{Qwen2-VL-7B-Instruct}~\cite{wang2024qwen2}, \textit{Pixtral-12B-2409}~\cite{agrawal2024pixtral}, and \textit{InternVL2-(2B, 4B, 8B)}~\cite{chen2024far}. The complete results are shown in Table~\ref{main results all}.

\subsection{Certain objects are more prone to hallucination.}
\label{appendix obj}
To gain deeper insights into the types of entities that are more prone to triggering hallucinations in MLLMs, we conducted a thorough statistical analysis of hallucination-associated entities produced by five representative MLLMs: \textit{Gemini-1.5-Flash}, \textit{Llama-3.2-11B-Vision-Instruct}, \textit{Qwen2-VL-(2B, 7B)-Instruct}, \textit{Pixtral-12B-2409}, and \textit{InternVL2-8B}. For each model, we systematically extracted and aggregated entities from the hallucinated outputs, allowing us to identify recurring patterns and quantify their frequency. Based on these findings, we established a comparative ranking of entities according to their tendency to induce hallucinations across different models.
The statistical results are shown in Figure~\ref{fig:obj}, objects such as ``table'', ``shirt'', and ``woman'' frequently appear in erroneous captions, indicating their potential role in inducing hallucinations. This is likely due to the inherent visual ambiguity or semantic complexity of objects, which makes them harder for models to accurately identify and categorize.

\subsection{Framework Adaptation Exploration Experiment Details}
\label{sec:Framework Adaptation Exploration}

We adopt the Unified Multimodal Hallucination Detection (\texttt{UniHD}) framework, which integrates \textit{Claim Extraction}, \textit{Autonomous Tool Selection}, \textit{Tool Execution}, and \textit{Hallucination Verification} for image-to-text and text-to-image hallucination detection. Using tools like object detection, attribute detection, scene text recognition, and commonsense knowledge search, the framework aims to improve performance on \texttt{CCHall} by bridging internal reasoning with real-world context.

Given that the \texttt{CCHall} focuses on image-to-text hallucinations, particularly object hallucinations, we adapt the \texttt{UniHD} framework to better suit the specific requirements of our benchmark. These adjustments are as follows:
\begin{enumerate}
    \item \textbf{\textit{Input Adaptation}}: For VQA tasks, we combine the question and answer into a declarative statement for clarity, e.g., ``What animal is in the box?'' and ``A bear'' become ``The animal in the box is a bear''. For Image Captioning, the input is the caption itself. Since \texttt{CCHall} is multilingual and some tools support only specific languages, all non-English inputs are translated to English for compatibility.
    \item \textbf{\textit{Claim Extraction and Query Generation}}: The adjusted framework begins by using \textit{Gemini-1.5-Flash} to perform claim extraction, breaking complete sentences into sub-claims for detailed hallucination analysis. For each extracted claim, \textit{Gemini-1.5-Flash} generates specific queries focused on object existence and commonsense verification, aligning with the object-focused nature of our benchmark.
    \item \textbf{\textit{Tool Execution}}: The framework automatically invokes \textit{Grounding DINO}~\cite{liu2024groundingdinomarryingdino} model to detect objects and return bounding box coordinates. Additionally, it utilizes the \textit{Serper Google Search API}\footnote{https://serper.dev} to perform internet-based fact verification by analyzing search results, comparing sources, and validating the truthfulness of each claim.
    \item \textbf{\textit{Integrated Evaluation}}: Evidence collected from the tools is integrated with the original image and its corresponding claims into a comprehensive prompt. \textit{Gemini-1.5-Flash} then evaluates each claim’s hallucination likelihood and provides reasoning, resulting in a final hallucination detection outcome.

\end{enumerate}

During experiments, the adapted \texttt{UniHD} framework executes 3,532 queries via the Serper API, averaging one query per call. As shown in Table~\ref{unihd appendix}, \texttt{UniHD} outperforms all other methods, achieving the best results in hallucination detection. These findings demonstrate the effectiveness of external tools and web-based verification for object existence validation, offering a clear advantage over traditional prompting and reasoning strategies. The results highlight the importance of integrating external resources to enhance hallucination detection, particularly for complex object hallucinations.

\subsection{Samples \& Error Analysis}
\label{sec:Error Analysis}
\begin{figure*}[t]
	\centering
	\includegraphics[width=\textwidth]{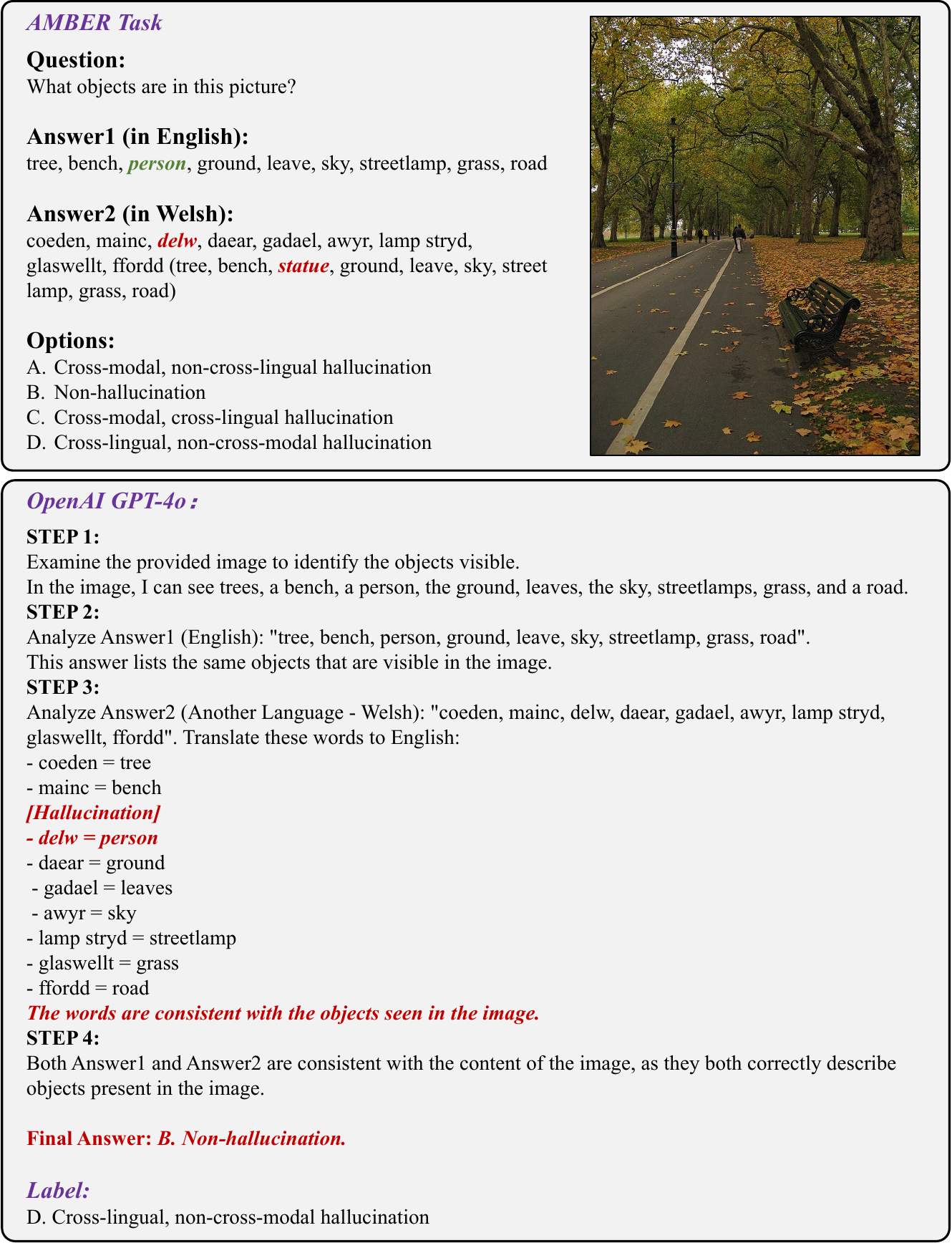}
	\caption{
    Error Analysis of Gemini on the AMBER}
	\label{fig:case_AMBER}
\end{figure*}

\begin{figure*}[t]
	\centering
	\includegraphics[width=\textwidth]{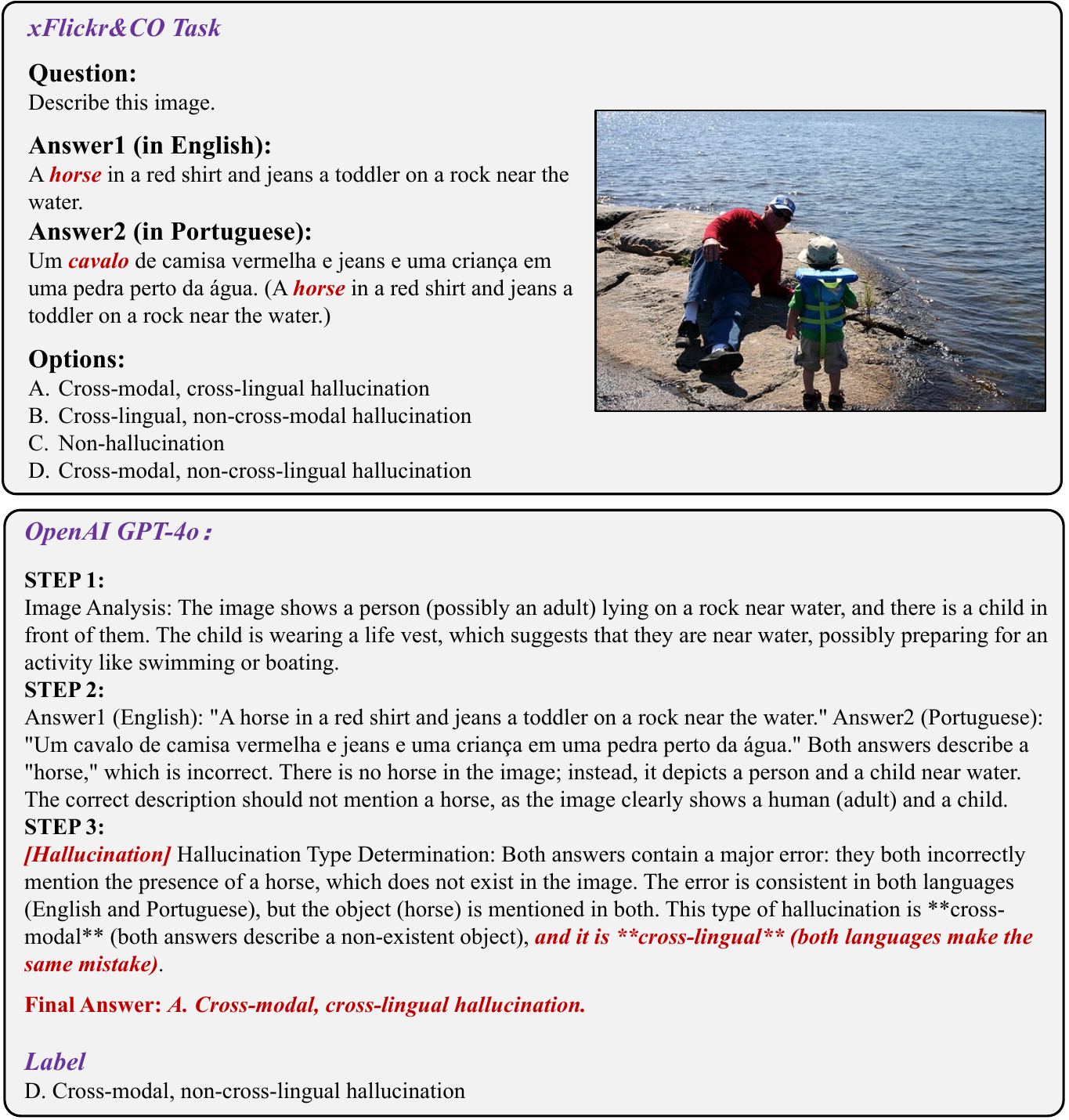}
	\caption{
    Error Analysis of Gemini on the xFlickr\&CO}
	\label{fig:case_xflickr}
\end{figure*}

\begin{figure*}[t]
	\centering
	\includegraphics[width=\textwidth]{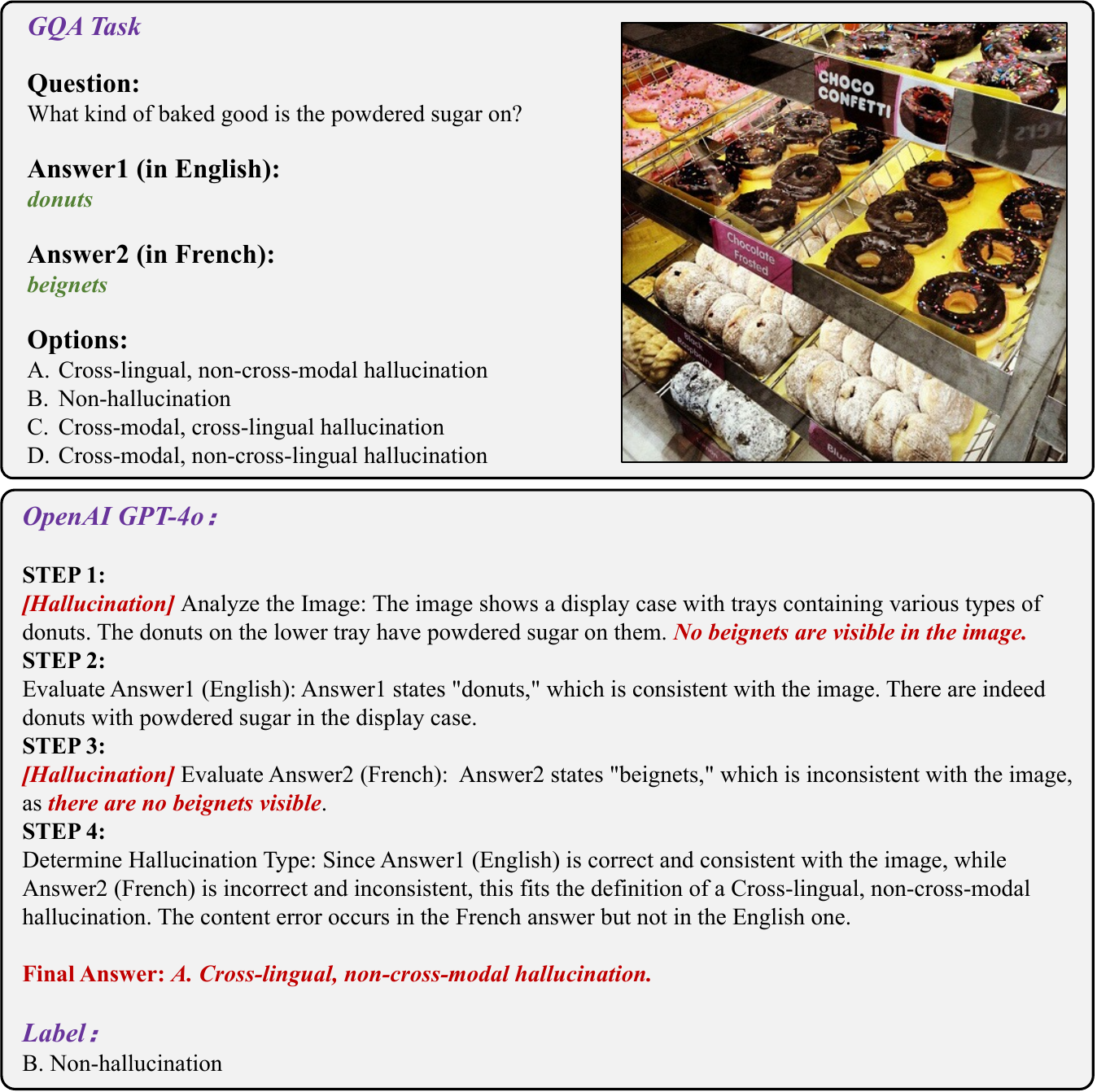}
	\caption{
    Error Analysis of Gemini on the GQA}
	\label{fig:case_gqa}
\end{figure*}

\begin{figure*}[t]
	\centering
	\includegraphics[width=\textwidth]{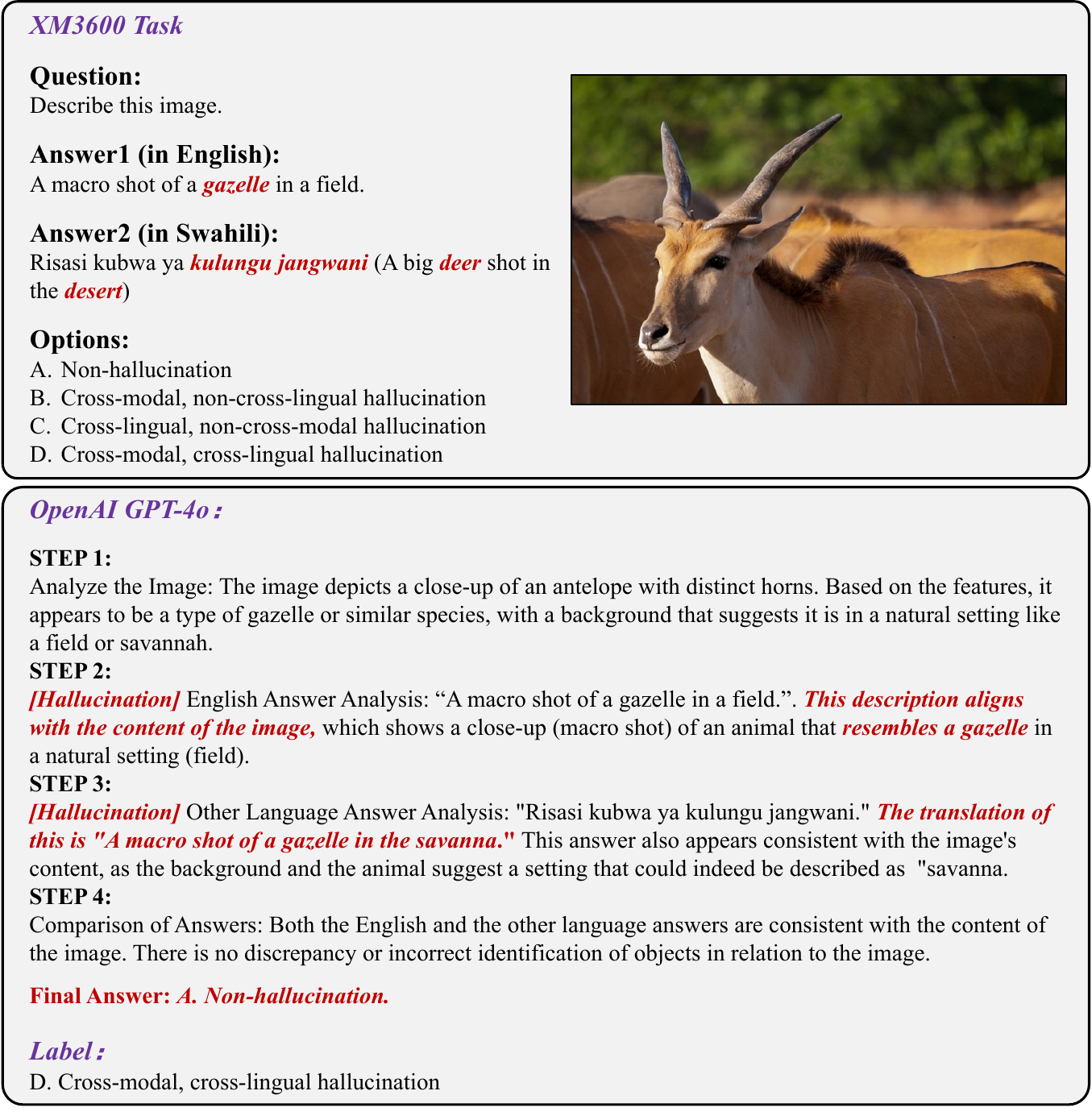}
	\caption{
    Error Analysis of Gemini on the XM3600}
	\label{fig:case_xm3600}
\end{figure*}

In this section, as shown in Figures~\ref{fig:case_AMBER}-\ref{fig:case_xm3600}, we conduct a detailed analysis of samples and error cases from the \textit{GPT-4o} across the four sub-datasets of the \texttt{CCHall} Benchmark: AMBER, xFlickrCO, GQA, and XM3600. For each of the four sub-datasets, we systematically collect and analyze error cases observed in \textit{GPT-4o}’s test-time performance. Through this detailed evaluation, we identify recurring failure patterns, domain-specific challenges, and systematic shortcomings. We aim to surface not only the nature of current limitations but also to quantify the actionable headroom.

\begin{table*}[t]
\setlength{\tabcolsep}{5pt}
\centering
\begin{adjustbox}{width=0.98\textwidth}
\begin{tabular}{lcccccccccc}
\toprule
\multirow{2}{*}{Model}  &   \multicolumn{2}{c}{\textbf{AMBER}} & \multicolumn{2}{c}{\textbf{GQA}} & \multicolumn{2}{c}{\textbf{xFlickr\&CO}} & \multicolumn{2}{c}{\textbf{XM3600}} & \multicolumn{2}{c}{\textbf{AVG}} \\
	\cmidrule{2-11}
           &  \textbf{Acc}  &  \textbf{Macro-F1}   & \textbf{Acc} & \textbf{Macro-F1}  &   \textbf{Acc} &  \textbf{Macro-F1}   &  \textbf{Acc} & \textbf{Macro-F1}  &  \textbf{Acc} & \textbf{Macro-F1}  \\
        \midrule
        \rowcolor[rgb]{ .980,  .980,  .980} 
        Random    & 25.1 & 30.0 & 25.0 & 29.4 & 24.9 & 30.3 & 25.1 & 29.9 & 25.0 & 29.9\\
		\midrule
		\rowcolor[rgb]{ .970, .978, .999 }
		\multicolumn{11}{c}{\rule{0pt}{2ex}\textit{\fontsize{11pt}{11pt}\selectfont Qwen2-VL-2B-Instruct} \cite{wang2024qwen2}}     \\  
		\midrule
		\rowcolor[rgb]{ .970, .978, .999 }
		\texttt{Direct} \cite{wang2024qwen2}  & 25.1 & 33.8 & 27.1 & 35.8  & 23.9 & 33.1  & 23.7 & 33.2  & 24.9 & 34.0\\
		\rowcolor[rgb]{ .970, .978, .999 }
		\texttt{CoT} \cite{kojima2022large}  & \underline{25.2} & 43.4 & \underline{27.2} & 35.0  & 24.1 & 52.1  & \underline{24.3} & \underline{52.5}  & \underline{25.2}  & \underline{45.7}\\
		\rowcolor[rgb]{ .970, .978, .999 }
		\texttt{SRO} \cite{lin2024interpreting}  & 24.7 & \underline{43.5} & 24.5 & 34.8  & \underline{24.6} & \underline{52.3}  & 24.2 & 52.4  & 24.5  & \underline{45.7}\\
		\rowcolor[rgb]{ .970, .978, .999 }
		\texttt{VDGD} \cite{ghosh2024vdgd}  & 23.7 & 36.3 & 19.8 & \underline{38.8}  & 20.2 & 44.2  & 23.7 & 41.9  & 21.8 & 40.3\\
		\rowcolor[rgb]{ .970, .978, .999 }
		\texttt{HalluciMAD}  \cite{lin2024interpreting} & 20.9 & 42.5 & 20.1 & 38.1  & 19.9 & 45.4  & 22.8 & 44.7  & 20.9 & 42.7 \\
		\midrule
        \rowcolor[rgb]{ .970, .978, .999 }
        \multicolumn{11}{c}{\rule{0pt}{2ex}\textit{\fontsize{11pt}{11pt}\selectfont InternVL2-2B} \cite{chen2024far}}     \\
		\midrule
		\rowcolor[rgb]{ .970, .978, .999 } 
		\texttt{Direct} \cite{chen2024far}  & \underline{24.4} & 28.1 & \underline{24.1} & 26.0  & \underline{24.6} & 26.9  & \underline{28.1} & 27.3  & \underline{25.3} & 27.1\\
		\rowcolor[rgb]{ .970, .978, .999 }
		\texttt{CoT} \cite{kojima2022large}  & 20.6 & 26.1 & 20.1 & 24.9  & 23.8 & 27.6  & 22.4 & 27.7  & 21.7 & 26.6\\
		\rowcolor[rgb]{ .970, .978, .999 }
		\texttt{SRO} \cite{lin2024interpreting}   & 9.7 & \underline{28.7} & 10.3 & \underline{28.9}  & 4.3   & \underline{28.4}  & \underline{28.1} & \underline{30.9}  & 8.9 & \underline{29.2}\\
		\rowcolor[rgb]{ .970, .978, .999 }
		\texttt{VDGD} \cite{ghosh2024vdgd}  & 14.9 & 25.0 & 18.4 & 28.1  & 16.3 & 27.1  & 18.4 & 26.9  & 17.0 & 26.8\\
		\rowcolor[rgb]{ .970, .978, .999 }
		\texttt{HalluciMAD}  \cite{lin2024interpreting} & 13.0 & 25.6 & 11.7 & 25.6  & 12.4 & 25.4  & 13.6 & 27.2  & 12.7 & 26.0\\
		\midrule
        \rowcolor[rgb]{ .970, .978, .999 } 
        \multicolumn{11}{c}{\rule{0pt}{2ex}\textit{\fontsize{11pt}{11pt}\selectfont InternVL2-4B} \cite{chen2024far}}     \\  
		\midrule
		\rowcolor[rgb]{ .970, .978, .999 }
		\texttt{Direct} \cite{chen2024far}  & \underline{25.0} & \underline{34.9} & \underline{25.2} & \underline{35.5}  & \underline{30.2} & 38.2  & \underline{30.2} & 38.3  & \underline{27.7}  & \underline{36.7} \\
		\rowcolor[rgb]{ .970, .978, .999 }
		\texttt{CoT} \cite{kojima2022large}  & 14.2 & 31.1 & 13.4 & 29.9  & 17.9 & 33.5  & 16.4 & 33.3  & 15.5 & 31.9\\
		\rowcolor[rgb]{ .970, .978, .999 }
		\texttt{SRO} \cite{lin2024interpreting}   & 12.3 & \underline{34.9} & 11.8 & 34.7  & 6.6   & 37.8  & 13.2 & \underline{38.9}  & 11.0 & 36.6\\
		\rowcolor[rgb]{ .970, .978, .999 }
		\texttt{VDGD} \cite{ghosh2024vdgd}  & 20.3 & 32.6 & 21.8 & 32.4  & 29.7 & \underline{39.1}  & 24.8 & 35.9  & 24.2 & 35.0\\
		\rowcolor[rgb]{ .970, .978, .999 }
		\texttt{HalluciMAD}  \cite{lin2024interpreting} & 16.7 & 31.9 & 16.2 & 33.0  & 21.1 & 37.0  & 29.4 & 34.9  & 20.8 & 34.2\\
		\midrule
        \rowcolor[rgb]{ .970, .978, .999 }
        \multicolumn{11}{c}{\rule{0pt}{2ex}\textit{\fontsize{11pt}{11pt}\selectfont InternVL2-8B} \cite{chen2024far}}     \\  
		\midrule
		\rowcolor[rgb]{ .970, .978, .999 }
		\texttt{Direct} \cite{chen2024far}  & 29.1 & 38.1 & 29.9 & 38.6  & 38.3 & 47.6  & 38.8 & 47.4  & 34.0 & 42.9\\
		\rowcolor[rgb]{ .970, .978, .999 }
		\texttt{CoT} \cite{kojima2022large}  & 31.3 & \underline{40.0} & \underline{33.6} & \underline{42.1}  & \underline{41.6} & 48.0  & \underline{40.1} & 47.6  & \underline{36.7} & \underline{44.4}\\
		\rowcolor[rgb]{ .970, .978, .999 }
		\texttt{SRO} \cite{lin2024interpreting}   & 30.3 & \underline{40.0} & 31.1 & 40.7  & 41.2 & \underline{48.6}  & 37.7 & 47.1  & 35.1 & 44.1\\
		\rowcolor[rgb]{ .970, .978, .999 }
		\texttt{VDGD} \cite{ghosh2024vdgd}  & \underline{33.2} & \underline{40.0} & 30.2 & 37.8  & 36.7 & 44.5  & 37.4 & 44.7  & 34.4 & 41.7\\
		\rowcolor[rgb]{ .970, .978, .999 }
		\texttt{HalluciMAD}  \cite{lin2024interpreting} & 29.9 & 38.6 & 30.0 & 39.0  & 37.9 & 45.9  & 39.6 & \underline{47.7}  & 34.3 & 42.8\\
		\midrule
        \rowcolor[rgb]{ .970, .978, .999 }
        \multicolumn{11}{c}{\rule{0pt}{2ex}\textit{\fontsize{11pt}{11pt}\selectfont Llama-3.2-11B-Vision-Instruct} \cite{meta2024llama}} \\
		\midrule
		\rowcolor[rgb]{ .970, .978, .999 }
		\texttt{Direct} \cite{meta2024llama}  & 31.6  & 38.8 & 32.1 & 38.9  & 35.4 & 43.2  & 43.3 & 49.4 & 35.6 & 42.6\\
		\rowcolor[rgb]{ .970, .978, .999 }
		\texttt{CoT} \cite{kojima2022large}  & 32.0  & 40.6 & 34.3 & 40.9  & 43.6 & 51.6  & \underline{46.4} & \underline{54.0} & 39.1 & \underline{46.8}\\
		\rowcolor[rgb]{ .970, .978, .999 }
		\texttt{SRO} \cite{lin2024interpreting}  & 32.1  & 40.6 & 34.4 & 40.9  & \underline{43.7} & \underline{51.7} & \underline{46.4} & \underline{54.0} & \underline{39.2} & \underline{46.8}\\
		\rowcolor[rgb]{ .970, .978, .999 }
		\texttt{VDGD} \cite{ghosh2024vdgd}  & \underline{34.0} & \underline{41.2} & \underline{35.6} & \underline{42.1}  & 36.6 & 45.4 & 42.4 & 50.7 & 37.1 & 44.9\\
		\rowcolor[rgb]{ .970, .978, .999 }
		\texttt{HalluciMAD}  \cite{lin2024interpreting} & 29.7 & 38.7 & 32.6 & 40.3  & 36.4 & 45.1  & 34.8 & 42.8 & 33.4 & 41.7\\
		\midrule
        \rowcolor[rgb]{ .970, .978, .999 }
        \multicolumn{11}{c}{\rule{0pt}{2ex}\textit{\fontsize{11pt}{11pt}\selectfont Qwen2-VL-7B-Instruct} \cite{wang2024qwen2}} \\
		\midrule
		\rowcolor[rgb]{ .970, .978, .999 }
		\texttt{Direct} \cite{wang2024qwen2}  & 36.2 & 39.5 & 33.3 & 38.3  & 42.9 & 46.7  & 39.9 & 44.4  & 38.1 & 42.2\\
		\rowcolor[rgb]{ .970, .978, .999 }
		\texttt{CoT} \cite{kojima2022large}  & 38.6 & 43.4 & 33.9 & \underline{39.0}  & 48.3 & 52.1  & \underline{48.4} & \underline{52.5}  & 42.3 & \underline{46.7}\\
		\rowcolor[rgb]{ .970, .978, .999 }
		\texttt{SRO} \cite{lin2024interpreting}  & \underline{38.7} & \underline{43.5} & 34.1 & 34.8  & \underline{48.4} & \underline{52.3}  & \underline{48.4} & 52.4  & \underline{42.4} & 45.7\\
		\rowcolor[rgb]{ .970, .978, .999 }
		\texttt{VDGD} \cite{ghosh2024vdgd}  & 34.8 & 36.3 & \underline{36.3} & 38.8  & 41.1 & 44.2  & 40.0 & 44.9  & 38.1 & 41.1\\
		\rowcolor[rgb]{ .970, .978, .999 }
		\texttt{HalluciMAD}  \cite{lin2024interpreting} & 37.0 & 42.5 & 31.4 & 38.1  & 38.0 & 45.4  & 38.6 & 44.7  & 36.3 & 42.7\\
		\midrule
        \rowcolor[rgb]{ .970, .978, .999 }
        \multicolumn{11}{c}{\rule{0pt}{2ex}\textit{\fontsize{11pt}{11pt}\selectfont Pixtral-12B-2409} \cite{agrawal2024pixtral}} \\
		\midrule
		\rowcolor[rgb]{ .970, .978, .999 }
		\texttt{Direct} \cite{agrawal2024pixtral}  & 23.9 & 38.0 & 34.8 & 45.5  & 43.3 & 51.3  & 38.0 & 48.9  & 35.0 & 45.9\\
		\rowcolor[rgb]{ .970, .978, .999 }
		\texttt{CoT} \cite{kojima2022large}  & 40.7 & 46.0 & 42.7 & 47.9  & 48.7 & 54.2  & 53.3 & 59.0  & 46.3 & 51.8\\
		\rowcolor[rgb]{ .970, .978, .999 }
		\texttt{SRO} \cite{lin2024interpreting}  & 43.0 & 47.8 & 41.3 & 48.1  & 50.6 & 55.5  & 51.6 & 57.8  & 46.6 & 52.3\\
		\rowcolor[rgb]{ .970, .978, .999 }
		\texttt{VDGD} \cite{ghosh2024vdgd}  & 43.4 & 43.1 & 45.1 & 47.3  & 47.3 & 48.4  & 56.4 & 58.2  & 48.1 & 49.3\\
		\rowcolor[rgb]{ .970, .978, .999 }
		\texttt{HalluciMAD}  \cite{lin2024interpreting} & \underline{46.3}  & \underline{49.8} & \underline{45.2}  & \underline{51.4}  & \underline{57.1}  & \underline{61.4}  & \underline{58.7}  & \underline{63.2}  & \underline{51.8} & \underline{56.4} \\
		\midrule
        \rowcolor[rgb]{ .961, .985, .970 }
        \multicolumn{11}{c}{\rule{0pt}{2ex}\textit{\fontsize{11pt}{11pt}\selectfont Gemini-1.5-Flash}  \cite{team2024gemini}} \\
		\midrule
		\rowcolor[rgb]{ .961, .985, .970 }
		\texttt{Direct} \cite{team2024gemini}  & 41.7  & 44.7 & 37.0 & 39.2 & 49.2 & 50.5 & 50.1 & 51.9 & 44.5 & 46.6 \\
		\rowcolor[rgb]{ .961, .985, .970 }
		\texttt{CoT} \cite{kojima2022large}  & 49.2 & 52.3 & 53.2 & 54.7 & 56.4 & 58.0 & 60.6 & 62.2 & 54.9 & 56.8 \\
		\rowcolor[rgb]{ .961, .985, .970 } 
		\texttt{SRO} \cite{lin2024interpreting}   & 49.0  & 52.9 & 51.6 & 53.4 & 55.7 & 58.3 & 58.2 & 60.8 & 53.6 & 56.4 \\
		\rowcolor[rgb]{ .961, .985, .970 }
		\texttt{VDGD} \cite{ghosh2024vdgd}  & \underline{58.9}  & \underline{60.3} & 50.6 & 52.8 & 52.7 & 54.0 & 60.6 & 62.4 & 55.7 & 57.4 \\
		\rowcolor[rgb]{ .961, .985, .970 }
		\texttt{HalluciMAD}  \cite{lin2024interpreting} & 52.2  & 54.5 & \underline{59.0} & \underline{60.6} & \underline{61.6}  & \underline{63.9} & \underline{63.7}  & \underline{64.9} & \underline{59.1} & \underline{61.0} \\
		\midrule
        \rowcolor[rgb]{ .961, .985, .970 }
        \multicolumn{11}{c}{\rule{0pt}{2ex}\textit{\fontsize{11pt}{11pt}\selectfont GPT-4o} \cite{achiam2023gpt}}  \\
		\midrule
		\rowcolor[rgb]{ .961, .985, .970 }
		\texttt{Direct} \cite{achiam2023gpt}  & 57.1  & 63.2 & 56.1  & 62.1 & 72.2  & 76.4 & 82.9  & 84.3 & 67.1  & 71.5 \\
		\rowcolor[rgb]{ .961, .985, .970 }
		\texttt{CoT} \cite{kojima2022large}  & 68.1  & 70.0 & 66.9  & 69.2 & 81.1  & 83.4 & 83.1  & 84.8 & 74.8  & 76.8 \\
		\rowcolor[rgb]{ .961, .985, .970 }
		\texttt{SRO} \cite{lin2024interpreting}  & 63.2  & 65.0 & 57.2  & 62.7 & 76.7  & 79.3 & 84.7  & 86.4 & 70.4  & 73.4 \\
		\rowcolor[rgb]{ .961, .985, .970 }
		\texttt{VDGD} \cite{ghosh2024vdgd}  & 64.7  & 66.6 & 56.3  & 62.5 & 72.3  & 77.0 & 83.2  & 85.6 & 69.1  & 73.0 \\
		\rowcolor[rgb]{ .961, .985, .970 }
		\texttt{HalluciMAD}  \cite{lin2024interpreting} & \textcolor{gold}{\ding{202}}  \textbf{70.9}  & \textcolor{gold}{\ding{202}}  \textbf{71.9}  & \textcolor{gold}{\ding{202}} \textbf{68.6}   & \textcolor{gold}{\ding{202}} \textbf{70.3}  & \textcolor{gold}{\ding{202}} \textbf{84.1}   & \textcolor{gold}{\ding{202}} \textbf{85.6}  & \textcolor{gold}{\ding{202}} \textbf{86.4}   & \textcolor{gold}{\ding{202}} \textbf{87.3}  & \textcolor{gold}{\ding{202}} \textbf{77.5}   & \textcolor{gold}{\ding{202}} \textbf{78.8}  \\
		\bottomrule
	\end{tabular}
	\end{adjustbox}
    \caption{The experimental results of Acc. (\%) and Macro-F1 score on MLLMs. The ``Random'' refers to the average performance obtained from three separate random selections. \raisebox{0.7mm}{\colorbox{myblue}{\textcolor{myblue}{\rule{0.7mm}{0.7mm}}}} represents the performance of open-source MLLMs, and \raisebox{0.7mm}{\colorbox{mygreen}{\textcolor{mygreen}{\rule{0.7mm}{0.7mm}}}} represents the performance of closed-source MLLMs. The \underline{underline} indicates better performance in the MLLM. \textcolor{gold}{\ding{202}}  represents the \textbf{Best} performance.}
\label{main results all}
\end{table*}

\end{document}